%% file: main.tex
\documentclass{article}

\usepackage[eandd,final,nonatbib]{neurips_2026}
\usepackage[numbers]{natbib}
\usepackage[utf8]{inputenc}
\usepackage[T1]{fontenc}
\usepackage{hyperref}
\usepackage{url}
\usepackage{booktabs}
\usepackage{amsfonts}
\usepackage{nicefrac}
\usepackage{microtype}
\usepackage[table]{xcolor}
\usepackage{graphicx}
\usepackage{amsmath}
\usepackage{subcaption}
\usepackage{longtable}
\usepackage{pdflscape}
\usepackage[title]{appendix}
\title{What DNA Foundation Models Learn Beyond Sequence Composition}
\author{%
  Vincenzo Yuto Civale \\
  University of Florence
  \And
  Andrew D. Bagdanov \\
  University of Florence
  \And
  Alberto Magi \\
  University of Florence
}

\begin{document}
\maketitle

\begin{abstract}
Do genomic foundation models (FMs) provide predictive information beyond simple sequence composition? We evaluate frozen representations from five DNA FMs against normalized $k$-mer frequencies across 57 genomic classification datasets and two exploratory gene-expression regression tasks. Simple $k$-mer features remain competitive with most models, whereas NTv3 shows a consistent advantage; the ranking is robust to globally fixed and multi-scale $k$-mer baselines and alternative downstream probes. To test incremental predictive utility directly, we compare global 6-mer features with their concatenation with each FM embedding under a matched Random Forest protocol. NTv3 improves 46 of 57 datasets (mean $\Delta$MCC $=+0.072$), while Caduceus provides smaller gains and the other models do not improve mean MCC. We separately decompose embeddings into a component linearly predictable from 6-mer frequencies and a ridge residual, finding that linear recoverability does not reliably indicate downstream utility. Exact 6-mer-preserving shuffles further reveal task-relevant sequence-order sensitivity, most pronounced for NTv3, although order sensitivity alone does not guarantee predictive improvements. Our results establish compositional features as essential controls and offer complementary diagnostics for evaluating the added value of frozen genomic representations, without making claims about fine-tuning capacity.
\end{abstract}


\section{Introduction}
DNA foundation models (FMs) have produced strong results in regulatory prediction, variant analysis, and sequence representation learning~\citep{boshar2025foundational,nguyen2023hyenadna,zhou2023dnabert,schiff2024caduceus,brixi2026genome}. Yet a practical question remains: \emph{when do their frozen representations provide predictive value beyond basic sequence composition?} Short motifs and nucleotide-content biases underlie many of the tasks used to assess these models. Normalized $k$-mer frequencies provide an inexpensive and interpretable control, but are not consistently included in FM evaluations~\citep{feng2025benchmarking,marin2023bend}.

Classical sequence representations have long been competitive, and prior work has already compared $k$-mer profiles with genomic foundation models in metagenomic binning~\citep{celikkanat2024revisiting}. Our aim is not to claim novelty for $k$-mer competitiveness itself. Instead, we assess supervised genomic tasks using a common frozen-feature protocol, test whether an FM adds value when explicitly combined with composition, and investigate which aspects of its embedding cannot be recovered linearly from the chosen compositional control.

\begin{figure}[t]
\centering
\begin{minipage}[c]{0.52\linewidth}\centering
\includegraphics[width=\linewidth]{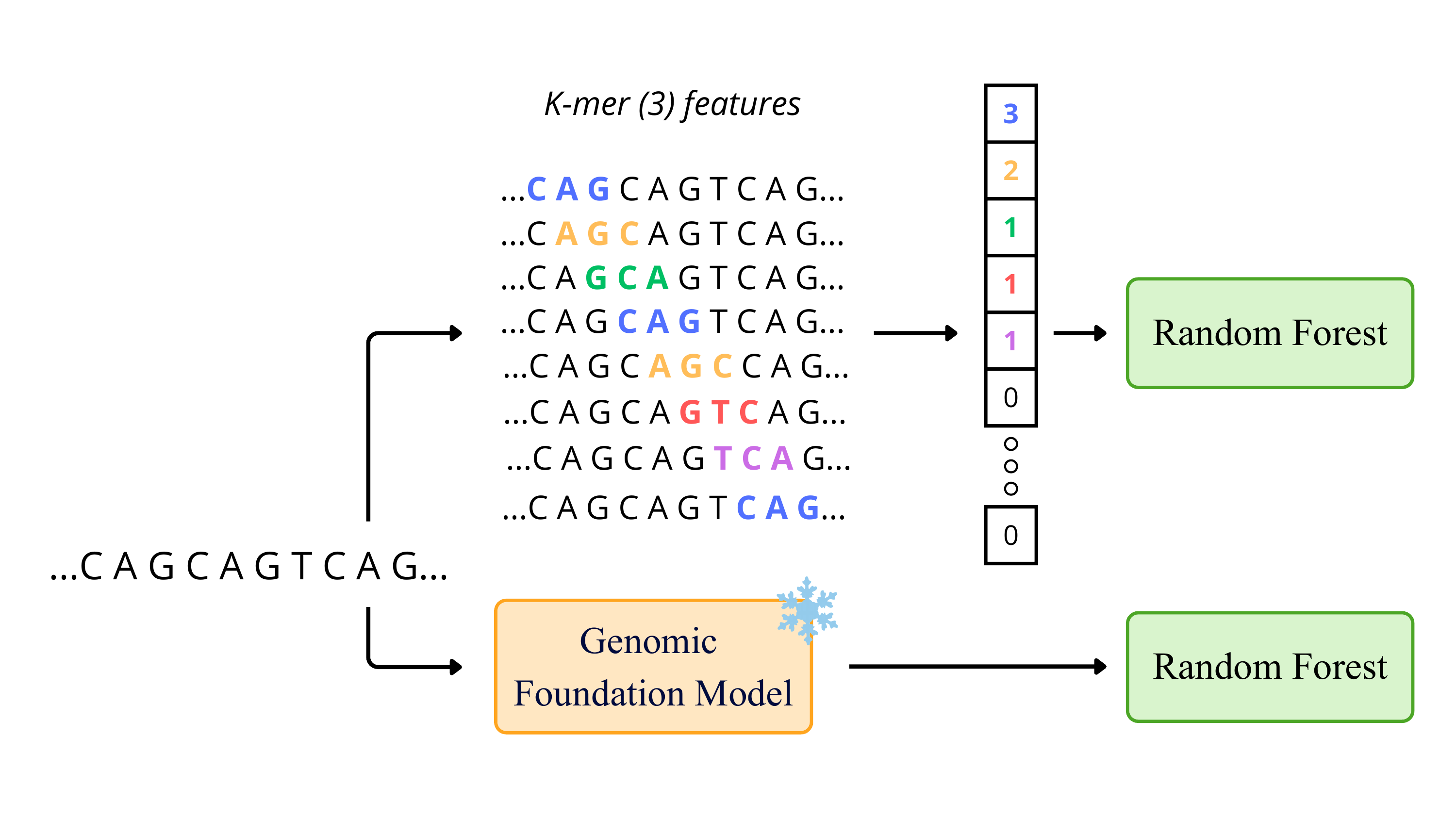}
\end{minipage}\hfill
\begin{minipage}[c]{0.45\linewidth}\centering
\includegraphics[width=\linewidth]{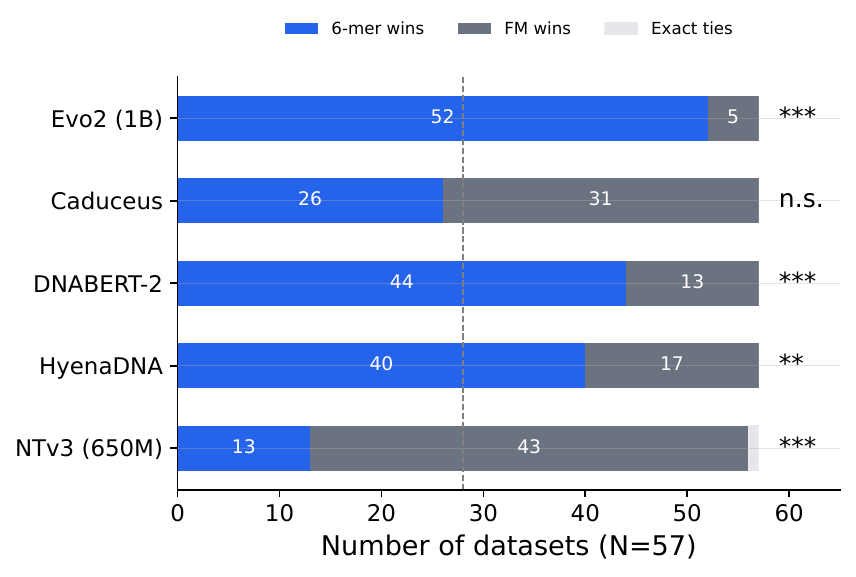}
\end{minipage}
\caption{
\textbf{Left:} Schematic illustration of $k$-mer
frequency extraction (shown for $k=3$ for visual
clarity) and frozen FM embedding extraction.
\textbf{Right:} Dataset-level wins and losses
across 57 classification datasets against a
globally fixed $k=6$ compositional baseline.
}
\label{fig:overview}
\end{figure}

We evaluate five architectures on 57 genomic classification datasets and report two gene-expression regression tasks as case studies. The classification analyses use a globally fixed $k=6$ reference, with fixed-$k$, multi-scale, and strand-aware checks. In addition to comparisons of individual feature spaces, we test the incremental utility of concatenating FM embeddings with 6-mer features. Finally, we use a ridge-based decomposition and exact composition-preserving sequence perturbations to assess linear recoverability and order sensitivity (Fig.~\ref{fig:overview}).

Our contributions are threefold:
\begin{enumerate}
\item A controlled comparison across 57 genomic classification tasks demonstrating when frozen DNA FM embeddings do, and do not, outperform fixed and multi-scale $k$-mer baselines; we check robustness to downstream probe and pooling choices.
\item A \emph{direct incremental-utility test} comparing 6-mer features with 6-mer--FM concatenations under matched classifiers, complemented by a diagnostic decomposition into linearly $k$-mer-predictable and unexplained embedding components.
\item A composition-preserving perturbation test showing that downstream-relevant order sensitivity is especially pronounced for NTv3, without equating unexplained embedding variation with a particular biological mechanism.
\end{enumerate}

\section{Related Work}
\label{sec:related}
\subsection{Genomic Foundation Models and Benchmarks}
\label{sec:related_fms}
Genomic FMs span transformer encoders, long-context implicit-convolution models, state-space architectures, and autoregressive sequence models~\citep{boshar2025foundational,nguyen2023hyenadna,zhou2023dnabert,schiff2024caduceus,brixi2026genome}. Broad evaluations such as the DNA Foundation Models Benchmark and BEND compare models across multiple genomic tasks but do not consistently employ explicit compositional controls~\citep{feng2025benchmarking,marin2023bend}. Our conclusions concern frozen, task-agnostic feature extraction; neither fine-tuning nor parameter-efficient adaptation is evaluated.

\subsection{Classical Sequence Features and Representation Diagnostics}
\label{sec:related_kmers}
Spectrum features and DNA sequence kernels have a long history~\citep{leslie2001spectrum}. They include mismatch-tolerant kernels~\citep{leslie2004mismatch}, gapped and position-aware variants, and gapped-$k$-mer methods for regulatory sequence analysis~\citep{ghandi2014gapped}. In particular, \citet{celikkanat2024revisiting} analyze scalable $k$-mer-profile representations for unsupervised metagenomic binning and compare them with genome FMs. We instead study supervised functional and regulatory classification across five frozen architectures and evaluate incremental predictive value beyond a specified global $k$-mer baseline. The normalized $k$-mer histogram is a \emph{minimal compositional control}, not an upper bound on classical sequence methods.

Our second diagnostic relates to representation probing~\citep{alain2016understanding,hewitt2019structural,tenney2019bert,belinkov2022probing}. Unlike a task-specific probe of a biological attribute, ridge regression predicts the FM embedding from a defined $k$-mer representation. The prediction error is a mapping-dependent residual; it must not be interpreted as inherently biological or non-compositional signal.

\section{Methods}
\label{sec:methods}
\subsection{Datasets and Features}
\label{sec:datasets}
We use 57 classification datasets from the DNA Foundation Models Benchmark~\citep{feng2025benchmarking}, covering promoters, enhancers, transcription-factor binding, splice sites, histone modifications, methylation-related sequence tasks, and other categories. Two tasks from the Genomics Long-range Benchmark~\citep{trop2024genomics}---Bulk RNA Expression and CAGE Prediction---were evaluated as exploratory regression case studies only; the Caduceus Bulk RNA result could not be certified against the official dataset configuration in the post-rebuttal audit, and is not used to support the paper's conclusions.

\subsection{Feature Representations}
\label{sec:representations}
For each input sequence, global $k$-mer frequencies are normalized counts of all contiguous subsequences of length $k$; the feature dimension is $4^k$. We use \emph{fixed $k=6$} as the primary baseline and assess fixed $k\in\{4,5,6\}$, their concatenation (multi-$k$), and reverse-complement-canonicalized features. We also retain the retrospective per-dataset best-$k$ oracle as a sensitivity analysis, \emph{not} as an actionable selection procedure. Frozen final-layer embeddings come from NTv3-650M, HyenaDNA, DNABERT-2, Caduceus-Ph, and Evo2-1B, with mean pooling as the primary token aggregation; alternative pooling and token choices are assessed separately.

\subsection{Evaluation and Incremental Predictive Utility}
\label{sec:protocol}
The standalone comparisons use the original benchmark Random Forest results. For incremental utility, the T6 protocol performs Random Forest model selection within the provided training split (200 or 500 trees; maximum depth 20; square-root feature subsampling; minimum split size 2), with stratified four-fold CV (shuffle and seed 42) and AUROC for binary or accuracy for multiclass tuning, followed by evaluation on the supplied held-out test split. Mean-pooled FM embeddings and L1-normalized 6-mer frequencies are used without scaling. The same split identifiers and reported test examples underlie the paired T6 contrasts; the historical standalone 6-mer scores are taken from the benchmark records, with reproducibility checked on a selected task. Because the baseline was not re-fitted independently on all 57 tasks in the supplied audit, the T6 protocol should be understood as a carefully aligned comparison rather than a fully independently re-executed baseline. The original benchmark, E3 decomposition, E4 probe robustness, E8 concatenation, and T6 experiments use related but non-identical pipelines; their effect sizes are therefore reported separately. All data-dependent preprocessing and model selection are restricted to training partitions. We do not construct new chromosome- or locus-disjoint splits; the source benchmark may therefore contain genomic overlap across folds. Matthews correlation coefficient (MCC) is the primary classification metric. Our robustness tests additionally use L2 logistic regression and a two-hidden-layer MLP (256 and 128 GELU units, dropout $0.1$, AdamW, early stopping; three seeds averaged for paired inference).

For a feature matrix $\mathbf K_6$ of L1-normalized 6-mer frequencies and an FM embedding matrix $\mathbf X$, the \emph{incremental-utility} comparison evaluates a classifier on $\mathbf K_6$ versus the same classifier on $[\mathbf K_6,\mathbf X]$. The T6 implementation uses mean-pooled FM embeddings, no feature scaler, and the fixed concatenation order [6-mer; FM]. Within T6 all splits, seeds, tuning criteria and training/test partitions are matched. A separately reported E8 analysis from the rebuttal used a broader maximum-depth grid and StandardScaler in its CV pipeline, so the numerical results of E8 and T6 are not interchangeable. We report dataset-level differences
\begin{equation}
\Delta_{\mathrm{add}}=\mathrm{MCC}([\mathbf K_6,\mathbf X])-\mathrm{MCC}(\mathbf K_6).
\label{eq:increment}
\end{equation}
A positive $\Delta_{\mathrm{add}}$ demonstrates incremental predictive utility in this \emph{specific} classifier and feature protocol. It does not prove that the FM represents information theoretically absent from 6-mer counts, or that the information cannot be recovered through a nonlinear transformation of those counts.

\subsection{Linear $k$-mer Recoverability and Ridge Residual}
\label{sec:decomposition}
For $n$ sequences, let $\mathbf K_6\in\mathbb R^{n\times4096}$ contain frequencies and $\mathbf X\in\mathbb R^{n\times d}$ the frozen embeddings. A ridge mapping is fit on the training fold to predict embeddings from 6-mer frequencies, with the same preprocessing applied to held-out examples:
\begin{align}
\hat{\mathbf X}&=f_{\mathrm{ridge}}(\mathbf K_6), \\
\mathbf E&=\mathbf X-\hat{\mathbf X}.
\label{eq:resid}
\end{align}
We call $\hat{\mathbf X}$ the \emph{linearly $k$-mer-predictable component} and $\mathbf E$ the \emph{ridge residual}. These terms are conditional on the feature basis, mapping, ridge penalty, and estimation procedure. In particular, ridge regression does \emph{not} in general produce an orthogonal decomposition, and the residual can include nonlinear compositional functions, order effects, noise, or modeling artifacts. For each embedding dimension $j$, we compute held-out $R_j^2=1-\sum_i(X_{ij}-\hat X_{ij})^2/\sum_i(X_{ij}-\bar X_{\mathrm{eval},j})^2$ and summarize $d^{-1}\sum_{j=1}^d R_j^2$ on held-out examples. This is an \emph{unweighted mean of per-dimension predictive $R^2$ values}, not necessarily the proportion of total embedding variance explained. We also evaluate $\hat{\mathbf X}$ and $\mathbf E$ separately as classifier features, as a secondary descriptive diagnostic rather than a causal attribution of predictive gains.

\subsection{Composition-Preserving Order Perturbation}
\label{sec:shuffle_method}
We use independently seeded Eulerian-path shuffles of a de Bruijn multigraph to preserve the exact global multiset of 6-mers for each sequence, following the general sequence-permutation construction of \citet{altschul1985permutation}. Three shuffled replicates are generated. Nontriviality and exact 6-mer conservation are verified; sequences without admissible nontrivial perturbations are retained unchanged and tracked. By construction, $\mathbf K_6$ is unchanged, and the prediction from a \emph{fixed} ridge map is therefore unchanged; separately refitted mappings may produce small changes. This intervention uses a \emph{linear probe}, not the benchmark Random Forest: a training-fitted scaler, fixed logistic-regression $C=1$, and ridge $\alpha=1$. We compare original and shuffled features using both an unchanged (frozen) probe and a retrained probe. Three shuffle seeds are averaged \emph{within each dataset} before the 57 paired dataset-level effects are summarized; the 171 dataset-by-seed observations are not treated as independent datasets. The intervention tests \emph{order sensitivity}; it does not distinguish absolute positions, relative motif arrangement, or a particular regulatory mechanism.

\section{Experimental Results}
\label{sec:results}
\subsection{Frozen Embedding Comparisons Across Compositional Baselines}
\label{sec:overall_comparison}
The original best-$k$ oracle comparison shows that NTv3 wins on 36 of 57 datasets (63.2\%), while HyenaDNA, DNABERT-2, and Evo2 lose on 49, 54, and 53 datasets, respectively (Appendix~\ref{app:statistical}). This oracle chooses $k$ retrospectively for each task and must not be read as a prospective baseline. Under globally fixed $k=6$, the corresponding number of datasets where $k$-mers outperform each model is 13 for NTv3, 40 for HyenaDNA, 44 for DNABERT-2, 26 for Caduceus, and 52 for Evo2 (Table~\ref{tab:fixedk}). The broad ranking is preserved across $k=4,5,6$ and multi-$k$ controls. Formal FDR-adjusted Wilcoxon and mixed-effects $p$ values in Appendix~\ref{app:statistical} refer to the \emph{retrospective best-$k$ oracle}, and must not be attributed to the fixed-6-mer comparison. NTv3 is the clearest frozen-embedding improvement; Caduceus is competitive, whereas the remaining three FMs frequently fall below compositional features.

\begin{table}[t]
\centering\small
\caption{Datasets (of 57) where the indicated compositional baseline outperforms the standalone frozen FM embedding. Best-$k$ denotes the retrospective task-specific oracle. These are \emph{loss counts for the FM}, not independent test-set significance estimates.}
\label{tab:fixedk}
\begin{tabular}{lrrrr}
\toprule
Model & Fixed $k=4$ & Fixed $k=5$ & Fixed $k=6$ & Best-$k$ oracle\\
\midrule
NTv3-650M & 16 & 16 & 13 & 20\\
HyenaDNA & 48 & 45 & 40 & 49\\
DNABERT-2 & 52 & 53 & 44 & 54\\
Caduceus-Ph & 27 & 30 & 26 & 32\\
Evo2-1B & 53 & 52 & 52 & 53\\
\bottomrule
\end{tabular}
\end{table}

\begin{figure}[t]
\centering\includegraphics[width=\linewidth]{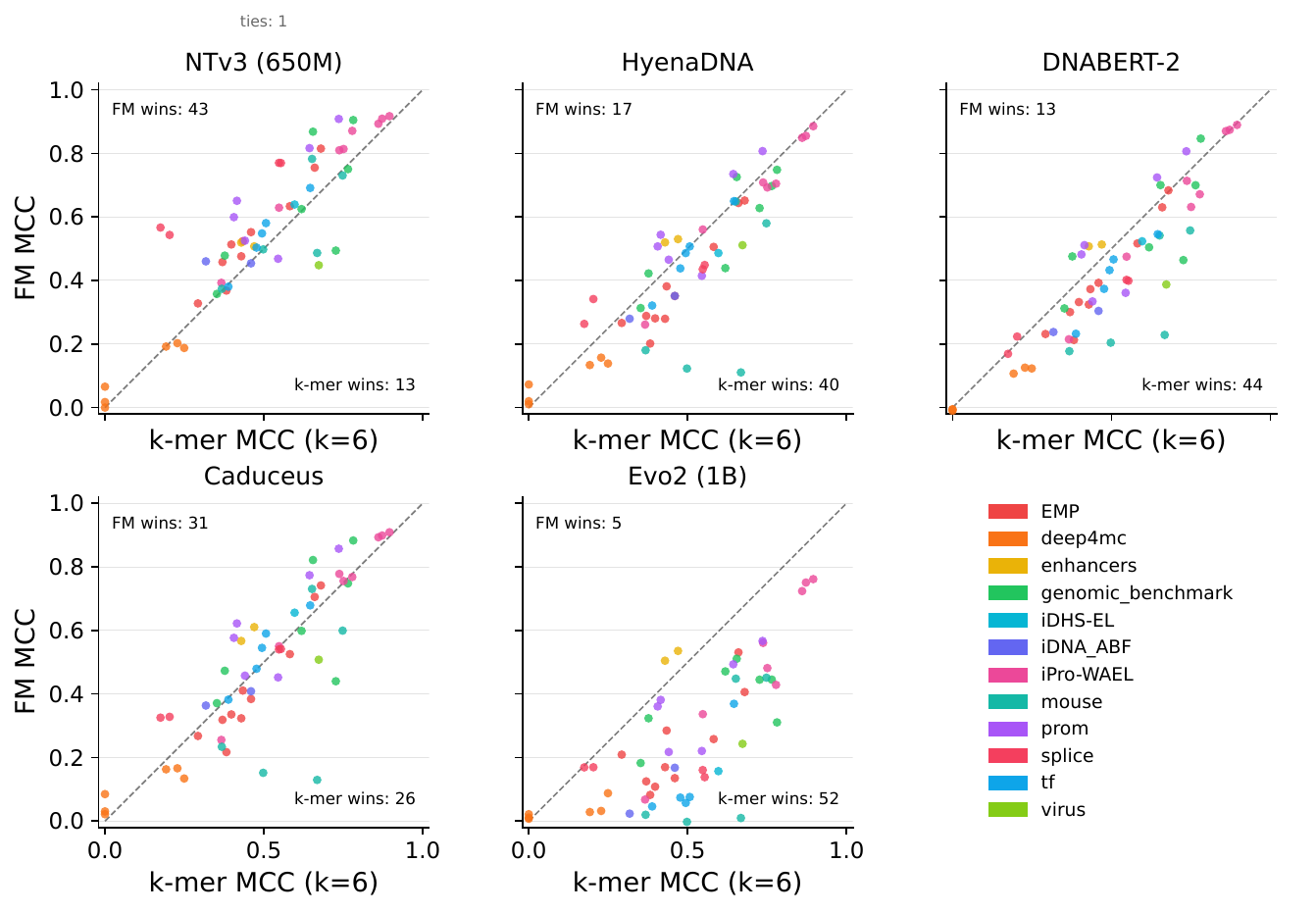}
\caption{Paired per-dataset MCC for each frozen embedding against the \emph{fixed 6-mer} Random Forest baseline. Each of the 57 datasets appears once per model, colored by benchmark category. The dashed diagonal marks equal MCC; model-level win/loss/tie counts refer to full-precision paired results. The corresponding best-$k$ oracle scatter is retained in the appendix.}
\label{fig:scatter_fms}
\end{figure}

\subsection{Added Value of FM Embeddings Beyond 6-mer Features}
\label{sec:incremental_results}
Concatenating an FM embedding with the global 6-mer vector and retraining the same Random Forest offers a more direct test of incremental predictive utility than analyzing ridge residuals alone. Under the T6 held-out benchmark protocol, NTv3 improves 46/57 tasks with a mean $\Delta_{\mathrm{add}}=+0.072$ MCC (median $+0.055$), and Caduceus improves 35/57 ($+0.022$; median $+0.012$). HyenaDNA ($-0.016$; 18/57 improved), DNABERT-2 ($-0.034$; 18/57), and Evo2 ($-0.075$; 8/57) do not increase mean performance. The E8 rebuttal protocol produces a closely related NTv3 result ($+0.071$, 45/57), which we report separately in Appendix~\ref{app:rebuttal_robustness}. The strongest NTv3 gains occur in splice, promoter, and other regulatory tasks, but this aggregate should not be taken as proof of a specific positional mechanism. Table~\ref{tab:t6_incremental} gives the complete model-level T6 summary; the figures show individual dataset-level effects.

\begin{table}[t]
\centering\small
\caption{Incremental utility in the T6 protocol, using exact 6-mer frequencies and frozen, mean-pooled FM embeddings. Differences are test MCC of [6-mer; FM] minus test MCC of 6-mer alone. Positive/negative/tie counts are based on unrounded dataset-level differences.}
\label{tab:t6_incremental}
\begin{tabular}{lrrrr}
\toprule
Model & Mean $\Delta$MCC & Median $\Delta$MCC & Improved & Worse / tied \\
\midrule
NTv3-650M & $+0.072$ & $+0.055$ & 46/57 & 11 / 0 \\
Caduceus-Ph & $+0.022$ & $+0.012$ & 35/57 & 22 / 0 \\
HyenaDNA & $-0.016$ & $-0.017$ & 18/57 & 39 / 0 \\
DNABERT-2 & $-0.034$ & $-0.042$ & 18/57 & 36 / 3 \\
Evo2-1B & $-0.075$ & $-0.065$ & 8/57 & 48 / 1 \\
\bottomrule
\end{tabular}
\end{table}

\begin{figure}[t]
\centering\includegraphics[width=\linewidth]{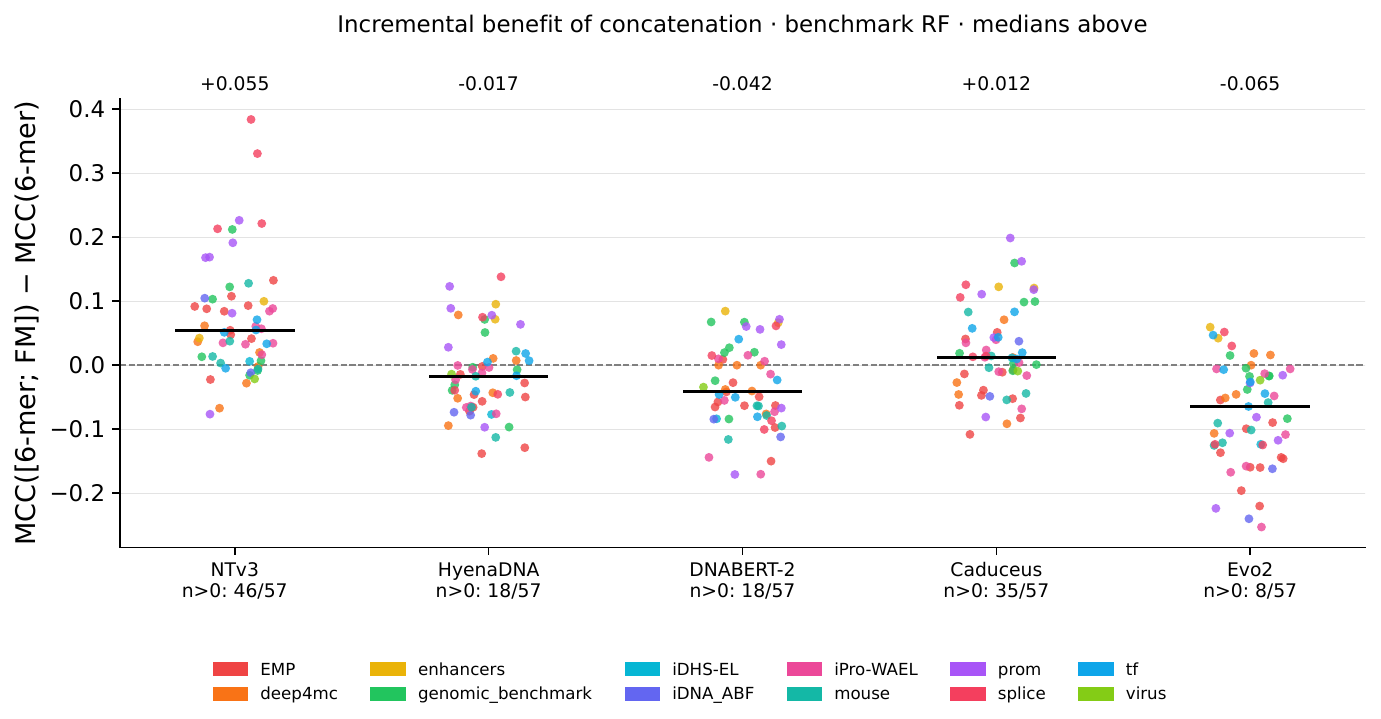}
\caption{T6 incremental predictive utility across 57 paired datasets per model, measured as $\mathrm{MCC}([6\mathrm{-mer},\mathrm{FM}])-\mathrm{MCC}(6\mathrm{-mer})$ on the held-out test set. Data use full-precision checkpoint metrics; reported summaries are rounded to three decimals. Paired differences correspond to the same dataset and fixed 6-mer reference. This plot uses the T6 Random Forest pipeline rather than the E8 rebuttal pipeline.}
\label{fig:incremental}
\end{figure}

\subsection{Linear Recoverability Is Not Predictive Utility}
\label{sec:decomposition_results}
\begin{figure}[t]
\centering\includegraphics[width=\linewidth]{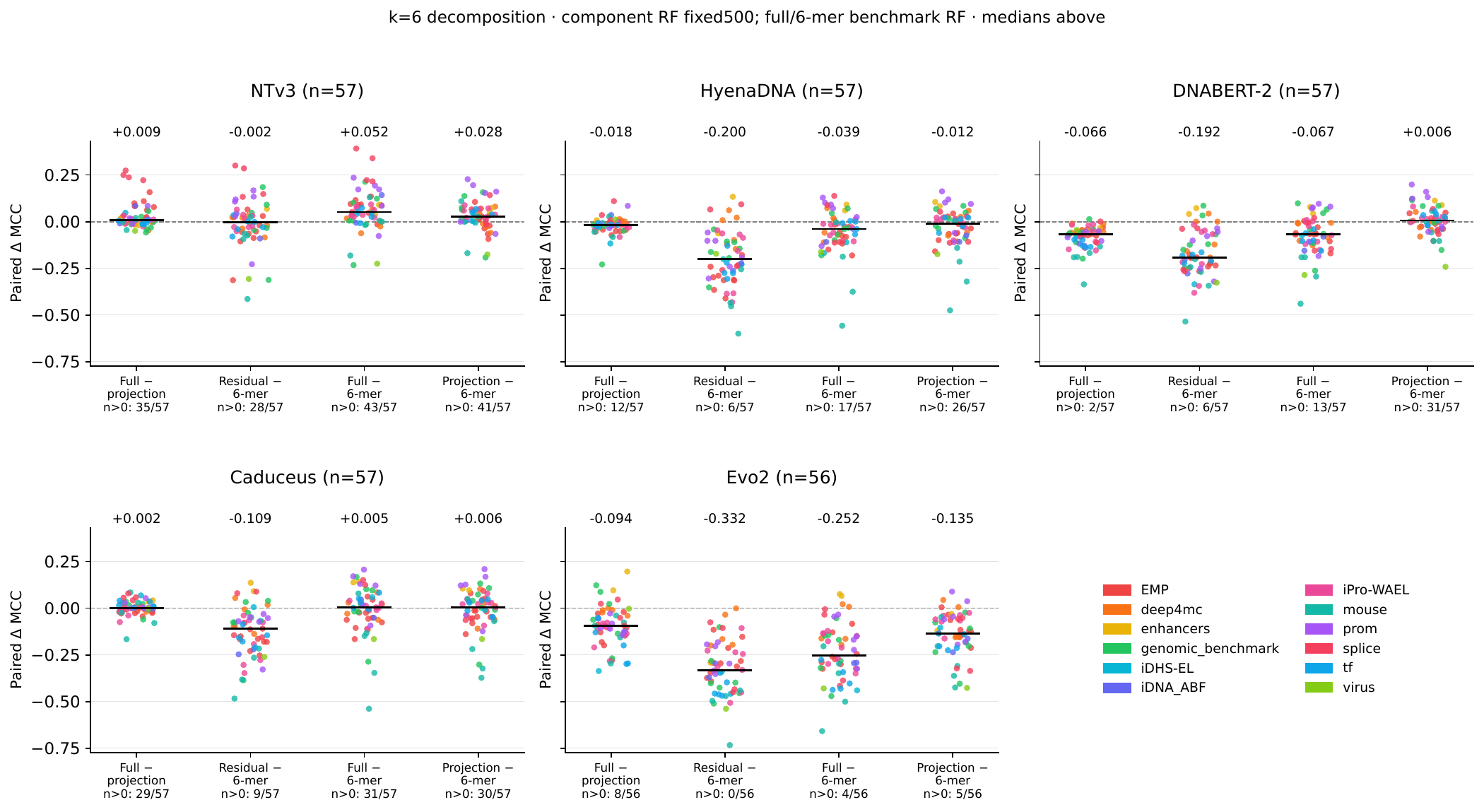}
\caption{Dataset-paired E3 MCC differences: full minus projection, residual minus fixed 6-mer, full minus fixed 6-mer, and projection minus fixed 6-mer (an explicit fourth diagnostic contrast). The component classifiers use an E3 fixed 500-tree Random Forest without CV or scaling, whereas the standalone full and 6-mer values come from the original RF benchmark. Thus this plot is a descriptive mixed-protocol diagnostic, not the matched E4 comparison. Evo2 includes 56 common datasets because one residual record is missing; the other models include 57. Medians and sign counts are computed before rounding.}
\label{fig:decomposition_boxplot}
\end{figure}

The median held-out dimension-averaged ridge $R^2$ is 0.75 for HyenaDNA, 0.71 for Caduceus, 0.51 for NTv3, 0.19 for DNABERT-2, and 0.17 for Evo2. This ordering does not track standalone predictive quality or the incremental gains from concatenation. In the E3 component analysis shown in Fig.~\ref{fig:decomposition_boxplot}, NTv3 has descriptive mean full-minus-prediction MCC $+0.030$ (median $+0.009$) across 57 paired tasks; the projection and full-embedding classifiers originate from the distinct E3 and benchmark procedures identified in the figure caption. In a separate E4 experiment with cross-validated probes and training-fold scaling, its corresponding mean differences are $+0.047$ for Random Forest, $+0.086$ for linear, and $+0.068$ for MLP; the E4 magnitudes must not be attributed to the E3 figure. In the E3 data, the ridge residual is weaker than the full embedding and its median MCC is slightly below the 6-mer baseline (median residual-minus-6-mer $=-0.002$); this does not contradict its carrying order-sensitive variation. For HyenaDNA, DNABERT-2, and Evo2 the predicted component is approximately as predictive as, or better than, the full embedding. Caduceus shows a smaller difference. These observations are \emph{diagnostic}: they do not establish that the ridge residual contains exclusively biological non-compositional content.

\subsection{Order Sensitivity Under Exact Composition Preservation}
\label{sec:order_results}
Global 6-mer counts are exactly invariant to our constrained sequence permutations, while FM embeddings may change. Across three shuffled replicas, the average MCC decreases with a frozen probe by 0.143 for NTv3, 0.063 for HyenaDNA, and 0.024 for DNABERT-2. Retraining the probe reduces these drops to 0.079, 0.030, and 0.013, respectively. The retrained NTv3--DNABERT-2 difference is 0.066 (95\% CI [0.044, 0.090]). NTv3 loses accuracy on 53/57 datasets after retraining (median drop 0.039; IQR 0.014--0.086). Its retrained-probe drop is associated with its standalone advantage over 6-mer features after adjusting for absolute performance (partial Spearman $\rho=0.430$).

\begin{figure}[t]
\centering\includegraphics[width=\linewidth]{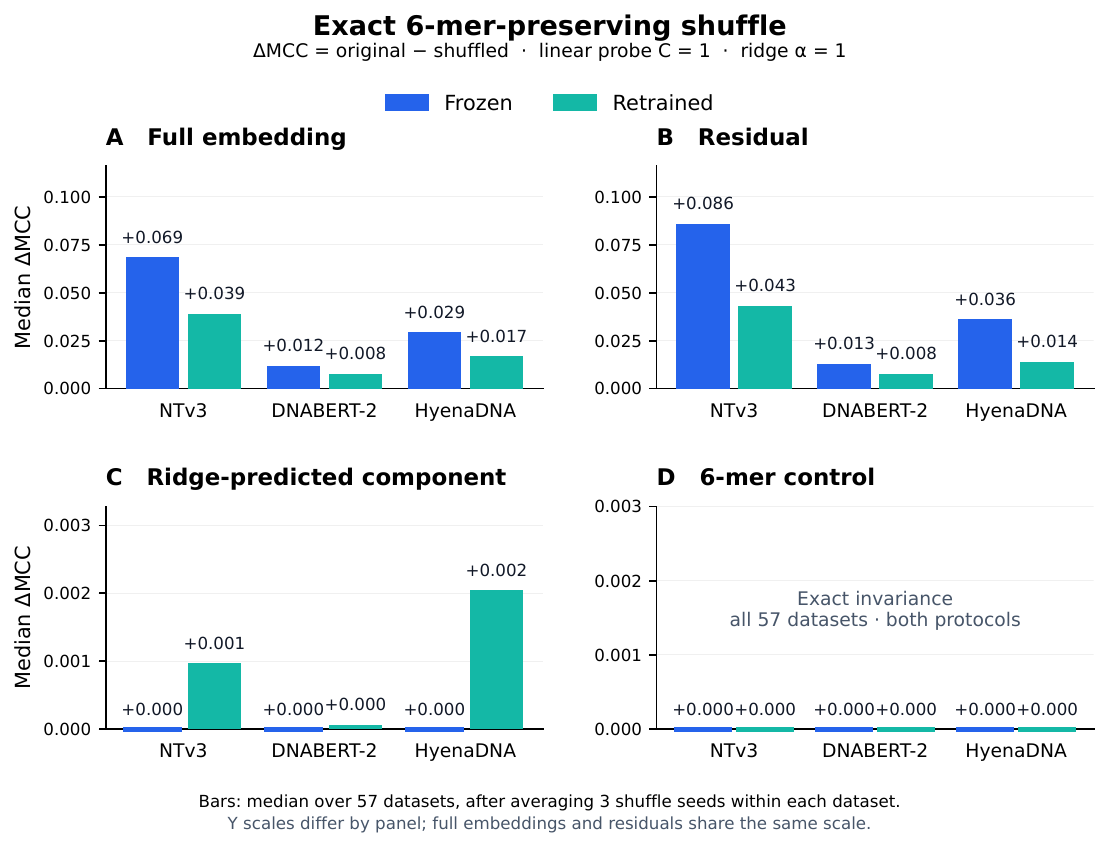}
\caption{MCC change (original minus shuffled) under exact preservation of global 6-mer counts, for full embeddings, ridge residuals, ridge-predicted components, and 6-mer controls, using a \emph{linear} downstream probe (training-fitted scaler, fixed $C=1$) and ridge $\alpha=1$. Three shuffle seeds are averaged within each of 57 datasets, and results are shown separately for frozen and retrained probes. The 6-mer input is invariant; small changes of the retrained predicted component arise from refitting rather than changed composition. No Caduceus or Evo2 perturbation results are claimed.
Bars report the median dataset-level MCC decrease
across 57 datasets after averaging three shuffle
seeds within each dataset. Mean decreases are
reported separately in Section 4.4 and Appendix A.
}
\label{fig:shuffle}
\end{figure}

This intervention gives evidence of downstream-relevant \emph{order-sensitive} variation, not a causal identification of particular motif arrangements or absolute genomic position. HyenaDNA also reacts to order perturbation but shows no mean incremental gain from concatenation; therefore order sensitivity is not sufficient for utility.

\subsection{Robustness to Probes, Pooling, and Strand Treatment}
\label{sec:robustness}
The standalone ranking is stable under Random Forest, L2 logistic regression, and MLP probes. Against fixed 6-mers, NTv3 achieves mean $\Delta$MCC $+0.061$, $+0.096$, and $+0.069$ under these probes, respectively; Caduceus remains approximately tied, while HyenaDNA, DNABERT-2, and Evo2 are below the reference (Appendix~\ref{app:rebuttal_robustness}). In particular, NTv3's largest advantage occurs under the linear probe, arguing against a simple Random-Forest-specific bias for sparse histograms.

Alternative token pooling does matter quantitatively: DNABERT-2 improves from mean MCC 0.428 (mean pooling) to 0.463 (CLS; 54 evaluable datasets), and Evo2 from 0.245 (mean) to 0.295 (max), without overturning the model-level conclusion. Canonical 6-mer features improve 20 of 26 prespecified strand-invariant datasets (median $+0.018$ MCC), whereas the median effect among 29 strand-specific tasks is null. The original, task-agnostic uncanonicalized 6-mer representation remains the uniform primary comparator; canonicalization is a biologically motivated sensitivity analysis.

\section{Discussion}
\subsection{Why Composition Is a Necessary Control}
\label{sec:discussion_kmers}
Genomic classifiers can exploit local motif frequencies and nucleotide-content differences even when sequence-order patterns are not modeled explicitly. The ability to match or exceed a frozen FM embedding with an inexpensive $k$-mer histogram is therefore an important evaluation outcome, not proof that the underlying biological task is determined by composition alone. Stronger classical features, including gapped and position-aware methods, can incorporate additional information; the 6-mer histogram should be viewed as a reproducible \emph{lower-complexity reference}, not the strongest conceivable non-neural baseline.

\subsection{What Does the Incremental Test Establish?}
\label{sec:discussion_cost}
The clearest positive result is the incremental utility of NTv3 for the tested frozen-feature Random Forest protocol: adding its representation to composition improves mean MCC and a large majority of tasks. Caduceus yields a smaller gain, while the other three models do not. Relative predictive gains should be considered alongside the substantial inference cost of FMs (Appendix~\ref{app:computational}), but utility depends on task and operating constraints. The significant perturbation response of HyenaDNA illustrates an important limitation of a purely representational description: the ability to encode order-dependent variation is not equivalent to adding predictive value beyond the selected composition features.

\subsection{Interpreting the Ridge-Based Diagnostic}
\label{sec:discussion_decomposition}
Ridge prediction identifies how well a \emph{specified linear map} from global 6-mer frequencies can recover each frozen representation. Its residual is neither orthogonal in general nor necessarily biological. A flexible nonlinear mapping might explain additional variation, and residuals can also reflect finite-sample error or artifacts. The fact that an embedding's mean per-dimension ridge $R^2$ is low does not mean that it contains a useful or predominantly order-specific feature space. Likewise, improved performance of the full embedding over its predicted component does not establish that the ridge residual will be an effective standalone representation. We therefore place the direct concatenation test ahead of the decomposition and treat the latter as an interpretable, conditional diagnostic.

\subsection{Scope and Limitations}
\label{sec:limitations}
This is a study of \emph{frozen} last-layer representations with the specified token pooling and probe protocols, not an upper bound on foundation models after full fine-tuning or parameter-efficient adaptation~\citep{hu2022lora}. We do not systematically probe intermediate layers. The datasets are inherited from a single benchmark, and their splits are not guaranteed to be locus- or chromosome-disjoint. Only two regression tasks are available, and the Caduceus Bulk RNA run could not be fully audited for reference assembly, TSS definition and effective context length; regression results are illustrative and do not justify general statements about genomic regression. Although robustness analyses cover multiple $k$ values, probes, pooling schemes, and strand conventions, richer nonlinear or position-aware compositional controls remain untested. The shuffling test shows order sensitivity but does not identify the biological mechanism or disentangle absolute location from relative motif organization. Finally, observed correlations with sequence length are not statistically significant after multiple testing and task-category adjustment (Appendix~\ref{app:rebuttal_robustness}); sequence length alone should not guide the choice of representation.

\paragraph{Broader Impacts.}
Our findings may improve the reliability and
computational efficiency of genomic foundation
model evaluation by encouraging the use of
inexpensive compositional baselines and more
transparent comparisons of predictive utility.
This may help reduce unnecessary computational
expenditure and support more reproducible
research. However, our conclusions could be
misinterpreted as evidence against foundation
models in settings not evaluated here, including
fine-tuning, variant interpretation, and clinical
prediction. Such extrapolations could lead to
inappropriate model selection or unsupported
biological conclusions. Our results should
therefore be interpreted strictly within the
evaluated frozen-representation setting.

\section{Conclusion}
Across 57 genomic classification tasks, normalized $k$-mer frequencies provide a strong baseline for frozen DNA foundation-model representations. A globally fixed 6-mer comparison and its robustness checks preserve the broad ranking, while a matched concatenation experiment demonstrates substantial incremental predictive utility for NTv3 and smaller gains for Caduceus. A ridge-based decomposition quantifies \emph{linear recoverability from the chosen features}, not intrinsically non-compositional biological signal. Exact composition-preserving perturbations add evidence that useful order-sensitive variation contributes to the observed NTv3 behavior, without claiming a unique mechanism. These results motivate compositional controls, matched incremental-utility tests, and careful scope restrictions in future evaluations of genomic FMs.

\section*{Funding}
This research received no specific funding or
external financial support.

\section*{Competing Interests}
The authors declare that they have no
competing interests.

\section*{Code and Data Availability}

The code and instructions required to reproduce
the experiments are publicly available at
\url{https://github.com/vincenzocivale/genome-cgr-embedding.git}.

The study uses publicly available benchmark
datasets and pretrained models. Dataset access
instructions and model identifiers are documented
in the repository.

\newpage

\bibliographystyle{plainnat}
\bibliography{references}

\newpage
\appendix

\begin{appendices}
\section{Additional Camera-Ready Analyses from the Rebuttal}
\label{app:rebuttal_robustness}

\subsection{Multi-scale Compositional Baseline}
Concatenating features for $k=4,5,6$ requires no per-task model selection. Compared with each standalone FM, the multi-$k$ representation has larger MCC on 19/57 NTv3 tasks (median multi-$k$ minus FM $=-0.025$), 47/57 HyenaDNA tasks ($+0.048$), 53/57 DNABERT-2 tasks ($+0.068$), 30/57 Caduceus tasks ($+0.002$), and 53/57 Evo2 tasks ($+0.241$). Retrospectively, the best single $k$ is 4 on 28 datasets, 5 on 13, and 6 on 16. A table listing best $k$ for each task requires the underlying task-level output rather than aggregate figures.

\subsection{Independent E8 Incremental-Utility Check}
The E8 experiment underlying the rebuttal uses $k=6$ and mean pooling, with a Random Forest grid of 200 or 500 trees and maximum depth 12 or 20, and scaling within the CV pipeline. It is an independent sensitivity analysis rather than the source of the T6 figure in the main paper.
\begin{table}[h]
\centering\small
\caption{E8 paired incremental MCC versus fixed global 6-mer frequencies; numbers are not pooled with the T6 results. All models have 57 paired datasets.}
\label{tab:e8_concat}
\begin{tabular}{lrrr}\toprule
Model & Mean $\Delta$MCC & Median $\Delta$MCC & Improved datasets\\\midrule
NTv3 & $+0.071$ & $+0.053$ & 45/57\\
Caduceus & $+0.022$ & $+0.010$ & 35/57\\
HyenaDNA & $-0.016$ & $-0.023$ & 18/57\\
DNABERT-2 & $-0.035$ & $-0.042$ & 18/57\\
Evo2 & $-0.076$ & $-0.056$ & 8/57\\
\bottomrule\end{tabular}
\end{table}

\subsection{Downstream Probe Robustness}
\begin{table}[h]
\centering\small
\caption{Mean paired MCC difference for standalone FM embedding minus \emph{fixed 6-mer} features (positive favors FM). Results from the rebuttal; these are not the retrospective best-$k$ contrasts in Table~\ref{tab:stats}.}
\label{tab:rebuttal_probes}
\begin{tabular}{lrrr}\toprule
Model & Random Forest & L2 linear & MLP\\\midrule
NTv3-650M & $+0.061$ & $+0.096$ & $+0.069$\\
HyenaDNA & $-0.052$ & $-0.009$ & $-0.016$\\
DNABERT-2 & $-0.088$ & $-0.078$ & $-0.089$\\
Caduceus-Ph & $-0.006$ & $+0.012$ & $+0.003$\\
Evo2-1B & $-0.222$ & $-0.260$ & $-0.260$\\
\bottomrule\end{tabular}
\end{table}

\subsection{Alternative Pooling and Token Representations}
\begin{table}[h]
\centering\scriptsize
\caption{Mean MCC under alternative token aggregation choices, as reported in the rebuttal. CLS and last-token columns are only available for the indicated models; these means must not be interpreted as matched 57-task comparisons when coverage differs.}
\label{tab:pooling_robustness}
\begin{tabular}{lrrrrrr}\toprule
Model & Mean & Max & Mean+max & CLS & Last & Attention\\\midrule
NTv3 & .602 & .565 & .601 & -- & -- & .441\\
HyenaDNA & .497 & .411 & .490 & -- & .331 & .491\\
DNABERT-2 & .428 & .381 & .423 & .463 & -- & .428\\
Caduceus & .518 & .465 & .522 & -- & .403 & .517\\
Evo2 & .245 & .295 & .279 & -- & -- & .245\\
\bottomrule\end{tabular}
\end{table}
For DNABERT-2, the CLS comparison is available on 54 datasets: it outperforms fixed 6-mer features on 19/54 tasks (mean $\Delta$MCC $=-0.047$), compared with 12/54 for mean pooling (mean $\Delta$MCC $=-0.082$). Different evaluation subsets explain why these tabulated means should not be substituted for the main benchmark's means without checking their provenance.

\subsection{Distribution of Embedding Recoverability}
\begin{figure}[h]
\centering\includegraphics[width=.80\linewidth]{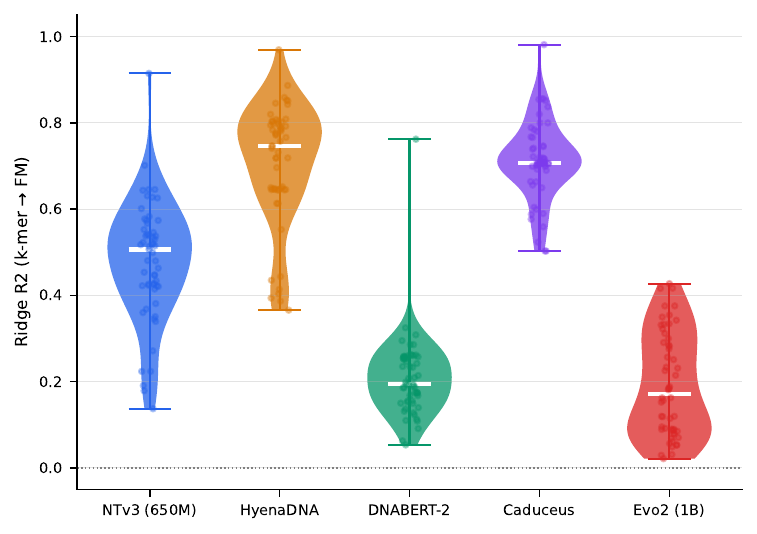}
\caption{Distribution of held-out average per-dimension ridge $R^2$ values for the fixed 6-mer predictor of each FM embedding. This measures linear predictability from the specified compositional basis; it does not quantify \emph{genuinely non-compositional information} or establish downstream relevance.}
\label{fig:violin_r2}
\end{figure}

\subsection{Residual Robustness and Order-Perturbation Details}
\begin{table}[h]
\centering\small
\caption{Median standalone ridge-residual MCC with a fixed 6-mer mapping versus a concatenated $k=4,5,6$ mapping. The residual is conditional on the mapping and is not a pure non-compositional feature.}
\label{tab:multik_residual}
\begin{tabular}{lrr}\toprule Model & Residual 6-mer & Residual multi-$k$\\\midrule
NTv3 & .487 & .486\\
HyenaDNA & .269 & .341\\
DNABERT-2 & .297 & .270\\
Caduceus & .336 & .373\\
Evo2 & .162 & .102\\
\bottomrule\end{tabular}
\end{table}
\begin{table}[h]
\centering\small
\caption{Mean MCC decreases under exact 6-mer-preserving shuffles; $\Delta$ is original minus shuffled performance. Bootstrap confidence intervals for the NTv3--DNABERT-2 difference are as reported in the rebuttal.}
\label{tab:shuffle}
\begin{tabular}{lrrrr}\toprule
Protocol & NTv3 & HyenaDNA & DNABERT-2 & NTv3--DNABERT-2 [95\% CI]\\\midrule
Frozen probe & .143 & .063 & .024 & .119 [.083, .160]\\
Retrained probe & .079 & .030 & .013 & .066 [.044, .090]\\
\bottomrule\end{tabular}
\end{table}
NTv3's ridge residual has a positive MCC decrease of 0.161 before retraining and 0.083 after retraining. The globally fixed 6-mer histogram is unchanged by construction. The perturbation does not establish which sequence-order attribute is responsible.

\subsection{Sequence Length Is Not an Explanatory Variable}
\begin{table}[h]\centering\small
\caption{Across 57 datasets, dataset-level Spearman correlation between sequence length and FM advantage over fixed 6-mer features, with Benjamini--Hochberg adjusted $p$ values. No association passes the specified multiple-test control.}
\label{tab:length_correlation}
\begin{tabular}{lrr}\toprule Model & Spearman $\rho$ & BH $p$\\\midrule
NTv3 & .235 & .390\\
HyenaDNA & -.106 & .721\\
DNABERT-2 & .143 & .721\\
Caduceus & -.049 & .895\\
Evo2 & -.001 & .996\\
\bottomrule\end{tabular}
\end{table}
The weak NTv3 trend also disappears after adjustment for task category, which is confounded with length. The regression tasks are too few to establish broad predictive trends.

\subsection{Original Oracle Scatter and Exploratory Regression Case Studies}
\begin{figure}[h]
\centering\includegraphics[width=\linewidth]{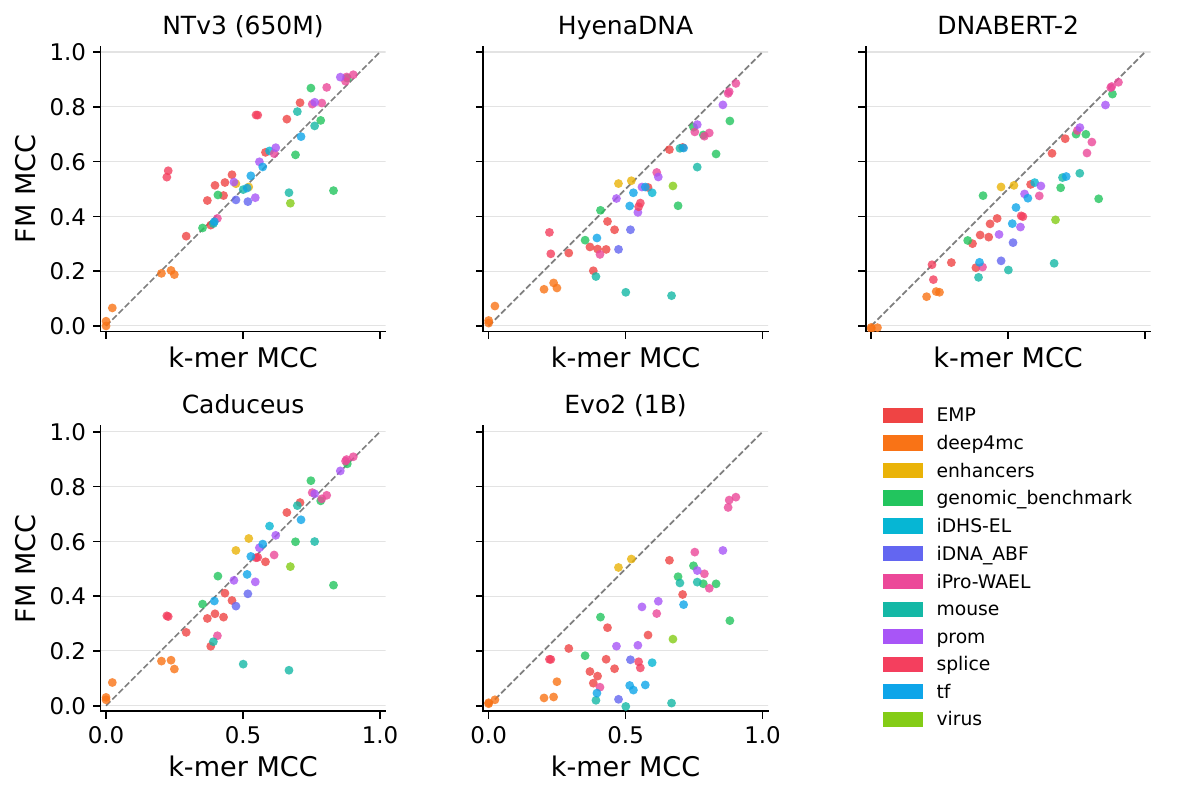}
\caption{Original classification comparison against the \emph{retrospectively chosen best $k$} on each dataset. This figure is retained for sensitivity comparison, not as the principal fixed-$k$ result.}
\label{fig:original_oracle_scatter}
\end{figure}
The two exploratory regression tasks (Bulk RNA expression and CAGE prediction)
were evaluated in the initial submission, but were not part of the 57-task
confirmatory classification analysis. We do not reproduce the historical
regression plot here, because the Caduceus Bulk RNA estimate could not be
independently audited against the benchmark's reference assembly, transcription
start site convention, and effective context length. The historical regression
analyses are not used for the model rankings or mechanistic conclusions in this
paper. A source-verified rerun would be necessary before interpreting the
unresolved Caduceus result quantitatively.

\section{Model Architectures}
\label{app:architectures}

We describe the five foundation models evaluated in this work, focusing on the
architectural and pre-training characteristics most relevant to interpreting the
decomposition results in Section~\ref{sec:decomposition_results}.

\paragraph{NTv3-650M~\citep{boshar2025foundational}.}
A hybrid U-Net/Transformer encoder with 650M parameters, pre-trained on a
multi-species corpus of approximately 9 trillion base pairs spanning 24 eukaryotic
species using a masked diffusion language modelling objective.
The model uses 6-mer-based tokenization. We do not assume that 6-mer tokenization alone guarantees preservation of biologically meaningful motif boundaries.
We extract mean-pooled embeddings over all non-padding tokens ($d = 1536$).
The maximum context window is 1 million base pairs.

\paragraph{HyenaDNA~\citep{nguyen2023hyenadna}.}
A decoder-based model using the Hyena operator, a data-controlled long
convolution, in place of self-attention, enabling sub-quadratic scaling with
sequence length.
With 1.6M parameters, it operates at single-nucleotide resolution using
character-level tokenisation (A/T/C/G), and is pre-trained exclusively on the
human reference genome (GRCh38, 3.2 Gbp) with a next-token prediction objective.
We use the HyenaDNA-160k variant for classification tasks ($d = 256$, context
window 160,000 bp) and HyenaDNA-1M for regression tasks ($d = 256$, context
window 1,048,576 bp).
Mean pooling is applied over all sequence positions.

\paragraph{DNABERT-2~\citep{zhou2023dnabert}.}
A BERT-architecture encoder with 117M parameters, using Byte Pair Encoding (BPE)
tokenisation trained on a multi-species corpus spanning 380 species, and ALiBi
positional embeddings for length generalisation.
BPE produces variable-length tokens and therefore differs from fixed 6-mer tokenization; we do not attribute observed performance differences to token-boundary effects without a controlled token-level analysis.
Pre-training uses a masked language modelling objective.
We extract mean-pooled embeddings ($d = 768$).

\paragraph{Caduceus-Ph~\citep{schiff2024caduceus}.}
A bidirectional, reverse-complement equivariant model based on the Mamba
state-space architecture (BiMamba blocks with parallel heads), with approximately
35M parameters.
It uses character-level tokenisation with reverse-complement augmentation, and is
pre-trained on the human reference genome with a masked language modelling
objective.
The maximum context window is approximately 131,000 bp.
We extract mean-pooled embeddings ($d = 256$).

\paragraph{Evo2-1B~\citep{brixi2026genome}.}
A large-scale autoregressive model based on the StripedHyena~2 architecture,
which combines input-dependent convolutions of varying ranges with sparse
attention layers.
With 1 billion parameters, it is pre-trained on OpenGenome2, a curated corpus of
approximately 8.8 trillion nucleotides spanning bacteria, archaea, eukaryotes, and
phages, using a next-token prediction objective at single-nucleotide resolution.
We extract mean-pooled embeddings from the final layer ($d = 1920$).

\section{Statistical Analyses}
\label{app:statistical}
\subsection{FM vs.\ K-mer: Wilcoxon and Mixed-Effects Results}
\label{app:wilcoxon}

We report formal statistical tests for the FM vs.\ best k-mer
comparisons across all 57 classification datasets. For each model,
$\Delta$ is defined as $\text{MCC}(\text{FM}) -
\text{MCC}(\text{best k-mer})$, so that $\Delta > 0$ indicates FM
advantage and $\Delta < 0$ indicates k-mer advantage.

\paragraph{Wilcoxon signed-rank test.}
For each model, we apply a two-sided Wilcoxon signed-rank test on the
57 paired $\Delta$ values to assess whether the distribution of
differences is systematically shifted from zero. $p$-values are
corrected for five simultaneous comparisons using the
Benjamini--Hochberg FDR procedure.

\paragraph{Mixed-effects model.}
To control for the substantial variability in task difficulty across
the benchmark, we additionally fit a linear mixed-effects model with
model identity as a fixed effect and a random intercept per dataset.
This isolates the systematic FM vs.\ k-mer difference from variation
attributable to dataset-level characteristics.

Results are reported in Table~\ref{tab:stats}.

\begin{table}[h]
\centering
\caption{Statistical comparison of FM vs.\ best k-mer MCC across
57 classification datasets. $\Delta = \text{MCC(FM)} -
\text{MCC(best k-mer)}$. Wilcoxon $p$-values are FDR-corrected
across five comparisons using the Benjamini--Hochberg procedure.
Mixed-effects estimates share the same $\Delta$ point estimate by
construction; the $p$-value reflects significance after absorbing
dataset-level variability via random intercepts.}
\label{tab:stats}
\smallskip
\resizebox{\linewidth}{!}{%
\begin{tabular}{lccccc}
\toprule
\textbf{Model}
  & \textbf{Mean $\Delta$}
  & \textbf{Median $\Delta$}
  & \textbf{Wins / Losses}
  & \textbf{Wilcoxon $p$ (FDR)}
  & \textbf{Mixed-effects $p$} \\
\midrule
NTv3-650M      & $+0.027$ & $+0.024$ & 36 / 20 & $0.007$               & $0.081$ \\
HyenaDNA-160k  & $-0.087$ & $-0.074$ & 8 / 49  & $1.90 \times 10^{-8}$ & $1.44 \times 10^{-8}$ \\
DNABERT-2-117M & $-0.108$ & $-0.095$ & 3 / 54  & $8.43\times 10^{-10}$ & $1.35 \times 10^{-12}$ \\
Caduceus-Ph    & $-0.042$ & $-0.013$ & 25 / 32 & $0.021$               & $0.006$ \\
Evo2-1B        & $-0.266$ & $-0.277$ & 4 / 53  & $5.39\times 10^{-10}$ & $1.62 \times 10^{-67}$ \\
\bottomrule
\end{tabular}
}
\end{table}

Under this retrospective best-$k$ oracle and the reported paired tests, the negative shifts for HyenaDNA, DNABERT-2, Caduceus, and Evo2 achieve the stated significance thresholds. These test statistics do not establish statistical significance for the separately evaluated fixed-6-mer baseline.
NTv3's positive shift reaches significance in the marginal Wilcoxon
test but is attenuated in the mixed model, indicating that its
advantage is task-selective rather than uniform, consistent with the category-level analysis in Section~\ref{sec:overall_comparison}, where NTv3 wins
concentrate in splice-site (100\%) and promoter (83\%) tasks and
drop to 0\% on virus and iDNA-ABF datasets.

Notably, Caduceus presents a dissociation between the two tests: the
Wilcoxon $p$-value (0.021) is marginal while the mixed-effects
estimate is more significant ($p = 0.006$). This reflects the fact
that Caduceus losses are more uniformly distributed across datasets, producing a consistent but small negative shift, whereas its
wins are concentrated in specific categories, inflating
within-dataset variance relative to the mixed-model residual.

\subsection{Classifier Choice: Random Forest vs.\ Linear Probe}
\label{app:classifier_ablation}

Our main evaluation uses a Random Forest (RF) classifier throughout, applied
symmetrically to all representations.
A natural alternative is a linear probe (LP), which directly tests whether
the information in a representation is linearly decodable, a property that
aligns with the linear structure of our geometric decomposition.
We compare RF and LP across all 57 classification datasets for each model and
for the k-mer baseline.

\begin{figure}[h]
    \centering
    \includegraphics[width=\linewidth]{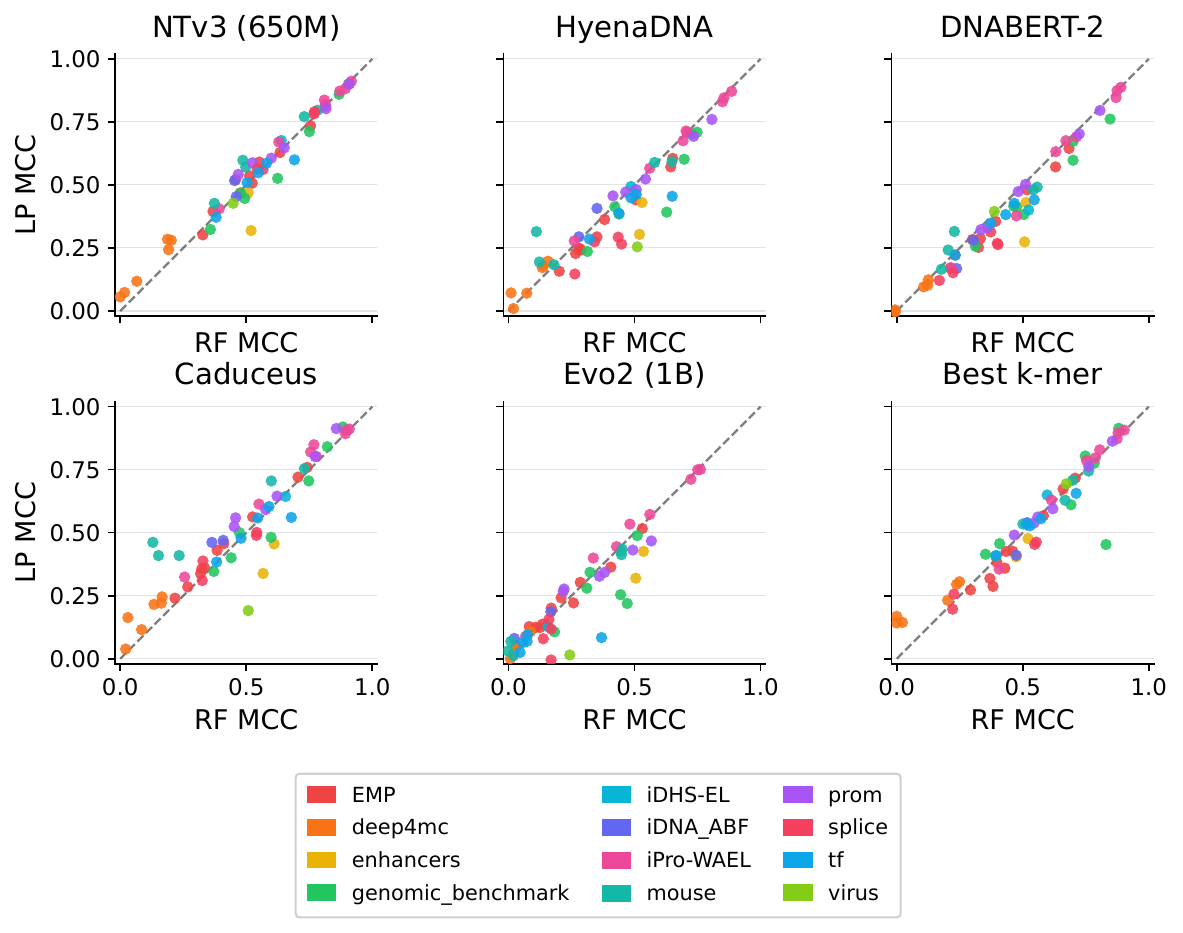}
    \caption{Linear probe (LP) MCC vs.\ Random Forest (RF) MCC across 57
    classification datasets for five foundation models and the best k-mer
    baseline, coloured by biological category.
    Points above the diagonal favour LP; points below favour RF.
    NTv3 and k-mers show no systematic advantage for either classifier.
    HyenaDNA and DNABERT-2 are significantly better under RF, while
    Caduceus benefits from LP.}
    \label{fig:rf_vs_lp}
\end{figure}

Figure~\ref{fig:rf_vs_lp} plots LP MCC against RF MCC for each method.
Points above the diagonal indicate LP advantage; points below indicate RF
advantage.

The results reveal a heterogeneous picture across models
(Table~\ref{tab:rf_vs_lp}).
In the historical probe comparison (not the E4 robustness matrix), NTv3 performs comparably under the two classifiers: LP outperforms RF in 25 of
57 datasets and RF outperforms LP in 14, with a non-significant mean difference
of $+0.009$ MCC ($p = 0.094$).
For k-mer features, the comparison is essentially symmetric (mean $\Delta =
-0.004$, $p = 0.908$), confirming that the choice of classifier does not
systematically favour or penalise compositional representations.

For HyenaDNA and DNABERT-2, RF is significantly better than LP: RF wins in
37 and 43 of 57 datasets respectively, with mean differences of $-0.036$
($p < 0.001$) and $-0.041$ ($p < 0.001$) in favour of RF.
This describes differences in linear and nonlinear probe behavior;
it does not identify the biological content or cause of the ridge residual.
Caduceus shows the opposite pattern: LP outperforms RF in 40 of 57 datasets
(mean $\Delta = +0.023$, $p = 0.002$), suggesting that its embedding geometry
is more amenable to linear classification.
Notably, this does not alter the main finding, even under LP, Caduceus
does not systematically outperform k-mer features.
Evo2 shows no significant difference between classifiers ($p = 0.121$).

\begin{table}[h]
    \centering
    \caption{Random Forest vs.\ linear probe comparison across 57 datasets.
    $\Delta = \text{LP} - \text{RF}$ MCC. Positive values favour LP;
    negative values favour RF.
    $p$-values from two-sided Wilcoxon signed-rank test.}
    \label{tab:rf_vs_lp}
    \small
    \begin{tabular}{lccccc}
        \toprule
        Model & LP $>$ RF & RF $>$ LP & Mean $\Delta$ & Median $\Delta$ & $p$-value \\
        \midrule
        NTv3-650M  & 25 & 14 & $+0.009$ & $+0.006$ & $0.094$ \\
        HyenaDNA   & 10 & 37 & $-0.036$ & $-0.032$ & $<0.001^{***}$ \\
        DNABERT-2  & 4  & 43 & $-0.041$ & $-0.036$ & $<0.001^{***}$ \\
        Caduceus   & 40 & 12 & $+0.023$ & $+0.023$ & $0.002^{**}$ \\
        Evo2-1B    & 21 & 26 & $-0.027$ & $-0.006$ & $0.121$ \\
        Best k-mer & 24 & 21 & $-0.004$ & $+0.004$ & $0.908$ \\
        \bottomrule
    \end{tabular}
\end{table}

These results support the use of RF as the primary classifier for two reasons.
First, it does not systematically disadvantage any representation: for NTv3 and
k-mers, the two methods that drive the main conclusions of the paper, the
classifier choice is inconsequential.
Second, RF avoids the assumption of linear separability, which is appropriate
given the heterogeneous geometry of the embeddings tested.
The significant LP advantage for Caduceus ($+0.023$ MCC) suggests that its
embeddings are more linearly structured than those of other models, a finding
consistent with its neutral decomposition profile (Section~\ref{sec:decomposition_results})
and worth investigating in future work.

\section{K-mer Baseline Analyses}
\label{app:kmer_analyses}

\subsection{$\boldsymbol{k}$-mer Selection by Biological Category}
\label{app:kselection}

\begin{figure}[h]
    \centering
    \includegraphics[width=\linewidth]{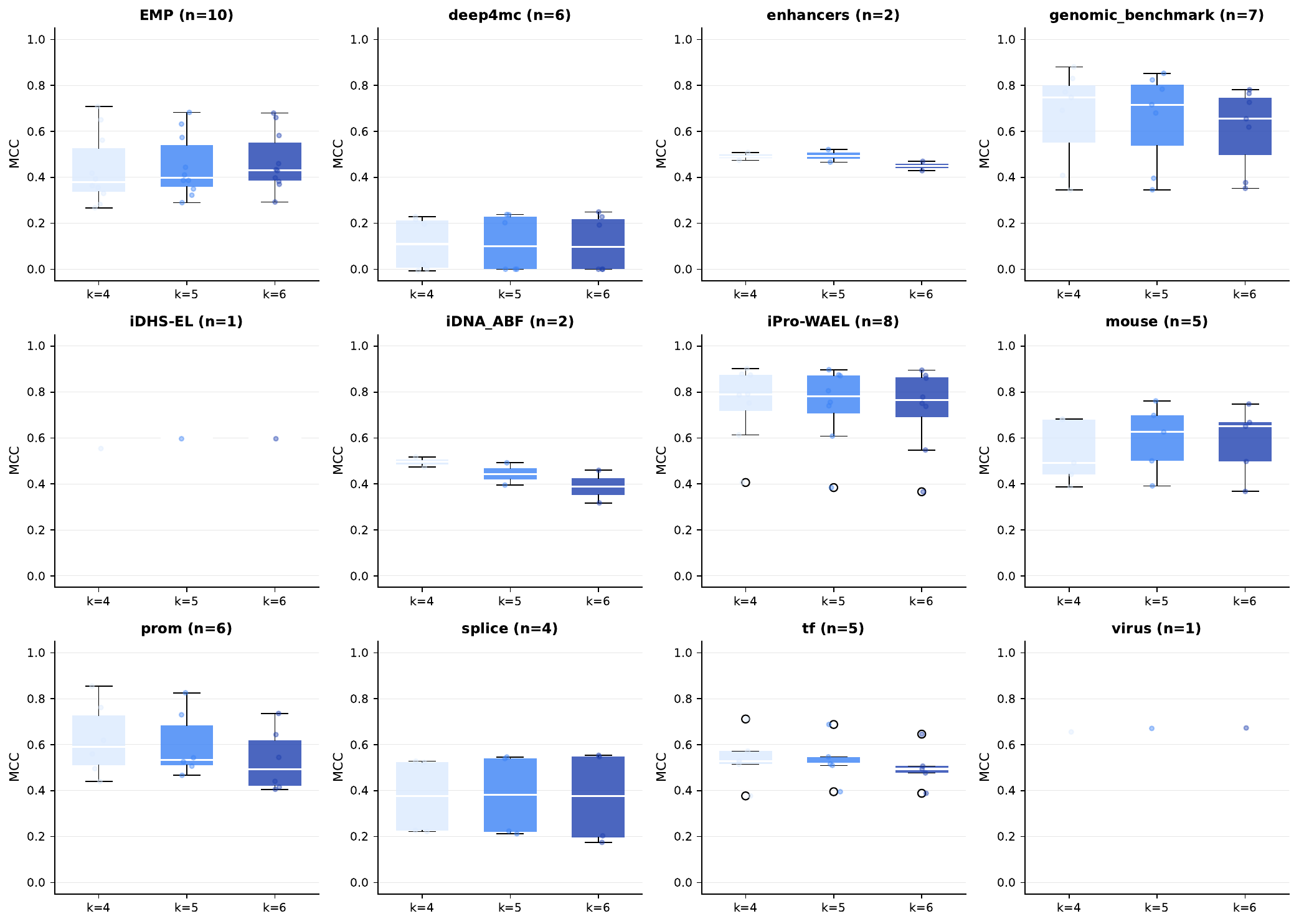}
    \caption{
    Distribution of Random Forest classification MCC
    across $k \in \{4,5,6\}$ for each biological task
    category, using the historical standalone $k$-mer
    benchmark results. Each box summarizes the
    distribution of dataset-level MCC scores within
    a category; the central line indicates the median,
    the box spans the interquartile range, and points
    outside the whiskers denote outliers.
    The optimal $k$ varies across tasks and categories.
    These descriptive results are distinct from the
    T6 incremental-utility experiment, which uses
    a globally fixed $k=6$ reference and four-fold
    cross-validation for hyperparameter selection.
    }   
    \label{fig:kmer_by_category}
\end{figure}

Figure~\ref{fig:kmer_by_category} shows the distribution of MCC across
$k \in \{4, 5, 6\}$ for each biological category. The optimal $k$ varies
systematically across categories: histone modification tasks (EMP) and
coding-sequence benchmarks favour $k = 6$, while methylation tasks
(iDNA\_ABF) show a reverse trend with $k = 4$ performing best, consistent
with the high GC content of CpG-dense regions. Promoter and enhancer
tasks are largely insensitive to $k$ selection. Splice-site tasks show
high within-category variance at all $k$, reflecting the positional
structure of splice signals that compositional features cannot fully
capture.

\subsection{Strand-Aware Canonical $\boldsymbol{k}$-mers}
\label{app:canonical_kmer}
For tasks where reverse-complement equivalence is biologically appropriate, we canonicalize a $k$-mer and its reverse complement into the same feature. At $k=6$, the number of features decreases from 4096 to 2080. Task context matters: we prespecified 26 strand-invariant tasks and 29 strand-specific tasks, excluding two ambiguous cases from this stratified comparison. Canonical 6-mers improve 20/26 strand-invariant tasks (median $+0.018$ MCC) but have zero median improvement among strand-specific tasks; the same qualitative pattern holds at $k=4$ and $k=5$. The pooled unstratified comparison in the original appendix obscured this relevant biological distinction. We retain ordinary non-canonicalized 6-mers as a fixed reference across all tasks rather than imposing strand symmetry where it may be invalid.\par
\begin{figure}[h]
\centering
\includegraphics[width=.80\linewidth]{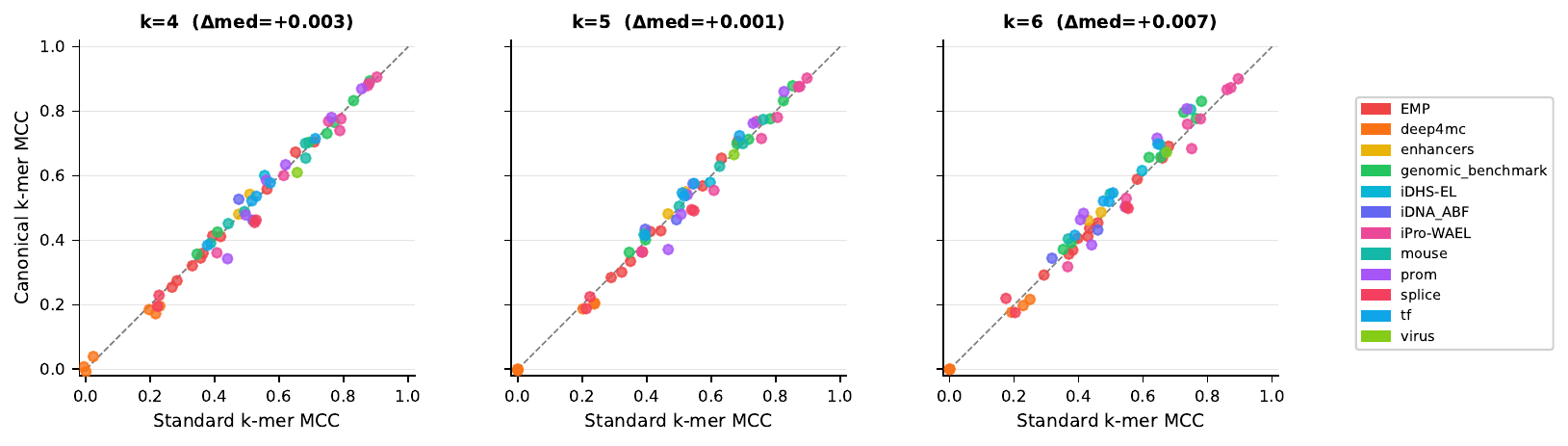}
\caption{Historical canonical-versus-standard $k$-mer comparison across the classification datasets. This pooled illustration does not replace the stratified analysis: reverse-complement canonicalization is biologically appropriate only for the prespecified strand-invariant tasks.}
\label{fig:canonical_kmer}
\end{figure}

\section{Per-Dataset Classification Results}
\label{app:full_results}

Table~\ref{tab:full_classification} reports MCC for all 57 classification datasets.
The \underline{underlined} value is the best $k$-mer baseline (selected across
$k \in \{4, 5, 6\}$); \textbf{bold} indicates the overall best result per dataset
across all methods. When the best $k$-mer is also the overall best, both styles apply.

\begin{landscape}
\begin{longtable}{rlcccccc}
\caption{Per-dataset MCC across 57 classification benchmarks. \underline{Underline}:
best $k$-mer baseline. \textbf{Bold}: overall best per dataset.
Category abbreviations: gb = genomic\_benchmark, iPro = iPro-WAEL.}
\label{tab:full_classification} \\

\toprule
\# & Dataset & Best $k$-mer & NTv3 & HyenaDNA & DNABERT-2 & Caduceus & Evo2 \\
\midrule
\endfirsthead

\toprule
\# & Dataset & Best $k$-mer & NTv3 & HyenaDNA & DNABERT-2 & Caduceus & Evo2 \\
\midrule
\endhead

\midrule
\multicolumn{8}{r}{\textit{Continued on next page}} \\
\endfoot

\bottomrule
\endlastfoot

1 & EMP/Yeast\_H3 & \underline{0.673} & \textbf{0.734} & 0.653 & 0.443 & 0.572 & 0.516 \\
2 & EMP/Yeast\_H3K14ac & \underline{0.383} & 0.536 & \textbf{0.544} & 0.352 & 0.288 & 0.124 \\
3 & EMP/Yeast\_H3K36me3 & \underline{0.426} & 0.591 & \textbf{0.594} & 0.419 & 0.356 & 0.137 \\
4 & EMP/Yeast\_H4 & \underline{0.716} & 0.812 & \textbf{0.827} & 0.717 & 0.645 & 0.363 \\
5 & EMP/Yeast\_H3K79me3 & \underline{0.567} & 0.628 & \textbf{0.656} & 0.550 & 0.481 & 0.221 \\
6 & EMP/Yeast\_H3K9ac & \underline{0.425} & 0.507 & \textbf{0.538} & 0.410 & 0.314 & 0.303 \\
7 & EMP/Yeast\_H3K4me2 & \underline{0.273} & 0.302 & \textbf{0.348} & 0.261 & 0.222 & 0.241 \\
8 & EMP/Yeast\_H4ac & \underline{0.359} & 0.468 & \textbf{0.495} & 0.332 & 0.252 & 0.201 \\
9 & EMP/Yeast\_H3K4me1 & \underline{0.318} & \textbf{0.521} & 0.505 & 0.321 & 0.283 & 0.123 \\
10 & EMP/Yeast\_H3K4me3 & \underline{0.287} & 0.395 & \textbf{0.400} & 0.220 & 0.170 & 0.128 \\
\midrule
11 & deep4mc/E.coli & \underline{0.141} & 0.056 & \textbf{0.181} & $-$0.007 & 0.000 & 0.000 \\
12 & deep4mc/G.pickeringii & \underline{0.144} & 0.118 & \textbf{0.148} & 0.012 & 0.004 & 0.022 \\
13 & deep4mc/G.subterraneus & \underline{\textbf{0.168}} & 0.073 & 0.103 & $-$0.003 & $-$0.004 & 0.038 \\
14 & deep4mc/C.elegans & \underline{0.305} & 0.285 & \textbf{0.353} & 0.119 & 0.188 & 0.108 \\
15 & deep4mc/D.melanogaster & \underline{0.295} & 0.281 & \textbf{0.357} & 0.135 & 0.246 & 0.052 \\
16 & deep4mc/A.thaliana & \underline{0.232} & 0.243 & \textbf{0.306} & 0.106 & 0.224 & 0.046 \\
\midrule
17 & enhancers/enhancer & \underline{\textbf{0.476}} & 0.468 & 0.425 & 0.446 & 0.430 & 0.426 \\
18 & enhancers/strength & \underline{\textbf{0.404}} & 0.319 & 0.344 & 0.254 & 0.274 & 0.319 \\
\midrule
19 & gb/enhancer\_cohn & \underline{0.456} & 0.467 & \textbf{0.499} & 0.456 & 0.379 & 0.343 \\
20 & gb/coding & \underline{0.804} & 0.859 & \textbf{0.880} & 0.742 & 0.673 & 0.488 \\
21 & gb/promoter\_notata & \underline{\textbf{0.775}} & 0.712 & 0.744 & 0.634 & 0.598 & 0.427 \\
22 & gb/human\_vs\_worm & \underline{0.914} & 0.899 & \textbf{0.925} & 0.845 & 0.761 & 0.280 \\
23 & gb/open\_chromatin & \underline{\textbf{0.414}} & 0.323 & 0.398 & 0.306 & 0.231 & 0.106 \\
24 & gb/enhancer\_ensembl & \underline{\textbf{0.611}} & 0.526 & 0.558 & 0.445 & 0.519 & 0.219 \\
25 & gb/regulatory\_region & \underline{0.453} & 0.446 & \textbf{0.516} & 0.476 & 0.334 & 0.254 \\
\midrule
26 & iDHS-EL/DNase\_I & \underline{0.649} & \textbf{0.676} & 0.665 & 0.581 & 0.401 & 0.126 \\
\midrule
27 & iDNA\_ABF/5mC & \underline{0.410} & 0.453 & \textbf{0.522} & 0.188 & 0.339 & 0.080 \\
28 & iDNA\_ABF/6mA & \underline{0.539} & 0.517 & \textbf{0.583} & 0.298 & 0.450 & 0.186 \\
\midrule
29 & iPro/Arab.\_TATA & \underline{0.796} & 0.820 & \textbf{0.840} & 0.747 & 0.632 & 0.534 \\
30 & iPro/Arab.\_NonTATA & \underline{0.829} & 0.872 & \textbf{0.885} & 0.767 & 0.675 & 0.445 \\
31 & iPro/B.amyloliq. & \underline{0.629} & \textbf{0.670} & 0.651 & 0.390 & 0.526 & 0.399 \\
32 & iPro/HUVEC & \underline{0.906} & \textbf{0.911} & 0.907 & 0.901 & 0.887 & 0.751 \\
33 & iPro/GM12878 & \underline{0.872} & \textbf{0.882} & 0.876 & 0.879 & 0.847 & 0.712 \\
34 & iPro/NHEK & \underline{0.786} & \textbf{0.837} & 0.761 & 0.729 & 0.691 & 0.572 \\
35 & iPro/R.capsulatus & \underline{0.355} & \textbf{0.405} & 0.334 & 0.150 & 0.332 & 0.088 \\
36 & iPro/Hela-S3 & \underline{0.895} & \textbf{0.901} & 0.891 & 0.884 & 0.874 & 0.749 \\
\midrule
37 & mouse/TFBS\_1 & \underline{0.534} & \textbf{0.571} & 0.364 & 0.249 & 0.334 & 0.031 \\
38 & mouse/TFBS\_2 & \underline{0.708} & \textbf{0.796} & 0.698 & 0.545 & 0.563 & 0.412 \\
39 & mouse/TFBS\_3 & \underline{0.744} & \textbf{0.771} & 0.683 & 0.565 & 0.491 & 0.434 \\
40 & mouse/TFBS\_4 & \underline{\textbf{0.628}} & 0.598 & 0.357 & 0.318 & 0.186 & 0.069 \\
41 & mouse/TFBS\_5 & \underline{0.408} & \textbf{0.427} & 0.372 & 0.197 & 0.217 & 0.012 \\
\midrule
42 & prom/tata\_300bps & \underline{0.490} & \textbf{0.588} & 0.547 & 0.354 & 0.323 & 0.267 \\
43 & prom/tata\_70bps & \underline{0.538} & 0.542 & 0.491 & 0.340 & \textbf{0.746} & 0.276 \\
44 & prom/all\_70bps & \underline{0.562} & \textbf{0.606} & 0.558 & 0.495 & 0.558 & 0.327 \\
45 & prom/all\_300bps & \underline{0.763} & \textbf{0.802} & 0.779 & 0.759 & 0.702 & 0.431 \\
46 & prom/notata\_70bps & \underline{0.595} & \textbf{0.647} & 0.604 & 0.538 & 0.585 & 0.343 \\
47 & prom/notata\_300bps & \underline{0.863} & \textbf{0.901} & 0.868 & 0.855 & 0.795 & 0.467 \\
\midrule
48 & splice/acceptors & \underline{0.453} & \textbf{0.790} & 0.414 & 0.392 & 0.274 & 0.156 \\
49 & splice/donors & \underline{0.462} & \textbf{0.782} & 0.412 & 0.384 & 0.352 & 0.079 \\
50 & splice/type\_NT & \underline{0.197} & \textbf{0.563} & 0.401 & 0.186 & 0.317 & 0.117 \\
51 & splice/type\_DNABERT & \underline{0.256} & \textbf{0.561} & 0.273 & 0.133 & 0.273 & $-$0.005 \\
\midrule
52 & tf/TFBS\_1 & \underline{0.555} & \textbf{0.586} & 0.559 & 0.454 & 0.495 & 0.095 \\
53 & tf/TFBS\_2 & \underline{0.527} & \textbf{0.548} & 0.519 & 0.429 & 0.474 & 0.064 \\
54 & tf/TFBS\_3 & \underline{\textbf{0.535}} & 0.510 & 0.488 & 0.381 & 0.348 & 0.069 \\
55 & tf/TFBS\_4 & \underline{\textbf{0.409}} & 0.372 & 0.357 & 0.265 & 0.221 & 0.025 \\
56 & tf/TFBS\_5 & \underline{\textbf{0.656}} & 0.600 & 0.531 & 0.454 & 0.441 & 0.084 \\
\midrule
57 & virus/covid\_variants & \underline{\textbf{0.694}} & 0.427 & 0.252 & 0.393 & 0.097 & 0.015 \\

\end{longtable}
\end{landscape}

\noindent\textit{Note.} This table reproduces the historical standalone
benchmark results against a retrospective best-$k$ oracle. These are not T6
concatenation outcomes. The primary paper's fixed 6-mer comparison is reported
separately in Fig.~\ref{fig:scatter_fms} and Table~\ref{tab:fixedk}.
Negative MCC values can occur for a classifier with poor predictions; their
presence alone does not diagnose a particular source of error.
Horizontal rules separate biological categories: EMP (1--10), deep4mc (11--16),
enhancers (17--18), genomic\_benchmark (19--25), iDHS-EL (26), iDNA\_ABF (27--28),
iPro-WAEL (29--36), mouse (37--41), prom (42--47), splice (48--51), tf (52--56),
virus (57).

\section{Computational Efficiency}
\label{app:computational}
\begin{figure}[h]
    \centering
    \includegraphics[width=0.75\linewidth]{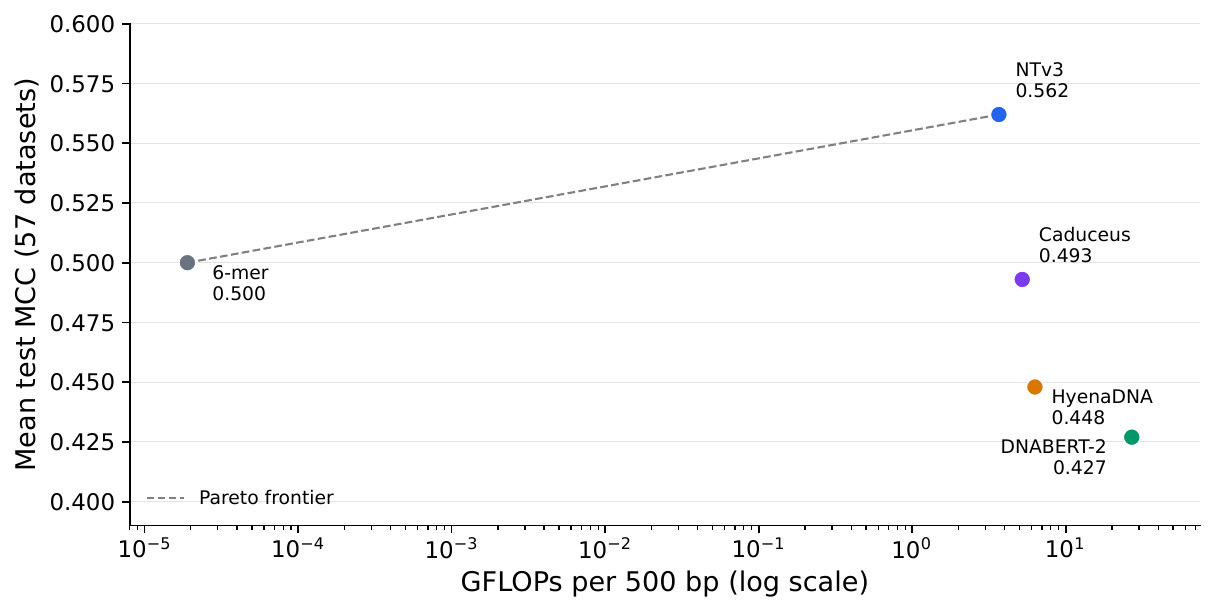}
    \caption{Mean classification MCC versus \emph{estimated GFLOPs per 500-bp sequence} (logarithmic cost axis). The fixed 6-mer feature vector is included at $1.9\times10^{-5}$ in the supplied cost data, although $k$-mer counting is not naturally a floating-point-operation workload. Evo2 is omitted because no verified corresponding 500-bp cost estimate is available. Costs are externally supplied estimates, rather than newly measured runtime or GPU throughput.}
    \label{fig:pareto}
\end{figure}

Figure~\ref{fig:pareto} shows the mean performance of the available fixed-6-mer benchmark features and a descriptive inference-cost comparison for the subset of models with supplied GFLOP estimates. The figure uses mean MCC 0.500 for the globally fixed 6-mer reference and 0.562 for NTv3; the original 0.535 k-mer MCC belongs to the retrospective best-$k$ oracle and must not be substituted here.

The supplied 500-bp computational-cost points are $1.9\times10^{-5}$ GFLOPs for 6-mer features, $3.67$ for NTv3, $5.21$ for Caduceus, $6.30$ for HyenaDNA, and $26.93$ for DNABERT-2. Their mean MCC values are 0.500, 0.562, 0.493, 0.448, and 0.427, respectively. Under these estimates, the 6-mer feature representation and NTv3 lie on the plotted cost-performance Pareto frontier; the other evaluated points are dominated. This is an illustrative complexity comparison rather than a hardware-normalized latency benchmark. Exact operation-count conventions, tokenization and sequence-length settings must be audited before claiming universal inference speedups.

\end{appendices}

\clearpage
\input{checklist}

\end{document}

%% file: checklist.tex
\section*{NeurIPS Paper Checklist}

\begin{enumerate}

\item \textbf{Claims}

Question: Do the main claims made in the abstract and introduction accurately
reflect the paper's contributions and scope?

Answer: \answerYes{}

Justification: The abstract and Introduction describe a controlled evaluation
of five frozen genomic foundation-model representations against global $k$-mer
frequencies on 57 classification datasets; a test of the incremental utility of
concatenating FM embeddings with fixed 6-mer features; and complementary
ridge-recoverability and exact 6-mer-preserving perturbation analyses. The
claims concern frozen representations under the specified downstream protocols,
not fine-tuned model capabilities. The two regression tasks are explicitly
presented as exploratory case studies. Sections~3--5 distinguish predictive
utility, linear recoverability, and sensitivity to sequence order without
claiming that ridge residuals are intrinsically biological or non-compositional.

\item \textbf{Limitations}

Question: Does the paper discuss the limitations of the work performed by the
authors?

Answer: \answerYes{}

Justification: Section~5.4 identifies the restriction to frozen final-layer
representations; lack of fine-tuning and systematic layer-wise evaluation;
possible genomic overlap inherited from benchmark splits; the limited scope of
two exploratory regression tasks; the unverified provenance of one historical
Caduceus Bulk RNA result; the conditional and linear nature of ridge
recoverability; the inability of the perturbation to identify a specific
biological mechanism; and the absence of more expressive, position-aware or
nonlinear compositional baselines. The observational analysis of sequence
length is also qualified.

\item \textbf{Theory assumptions and proofs}

Question: For each theoretical result, does the paper provide the full set of
assumptions and a complete (and correct) proof?

Answer: \answerNA{}

Justification: The paper presents an empirical
evaluation framework and does not introduce
formal theoretical results, theorems, or lemmas
requiring mathematical proofs. The ridge-based
decomposition in Section~3.4 is a methodological
diagnostic whose assumptions, interpretation,
and limitations are explicitly discussed.

\item \textbf{Experimental result reproducibility}

Question: Does the paper fully disclose all the information needed to reproduce
the main experimental results of the paper to the extent that it affects the
main claims and/or conclusions of the paper (regardless of whether the code and
data are provided or not)?

Answer: \answerYes{}

Justification: Sections~3.1--3.5 identify the benchmark datasets, provided
train/test partitions, frozen model variants, 6-mer representation, classifier
selection, ridge mapping, and composition-preserving perturbation. The primary
T6 incremental-utility analysis uses the held-out benchmark test sets and
training-only four-fold stratified cross-validation (seed~42) for Random Forest
selection over 200 or 500 trees, with maximum depth~20. The E8 concatenation,
E3 component, and E4 probe analyses have related but non-identical pipelines
and are reported separately rather than numerically pooled. Shuffle results
use a linear probe with fixed $C=1$, ridge $\alpha=1$, and three perturbation
seeds averaged within each dataset. The paper explicitly discloses that the
historical standalone 6-mer baseline was not independently refitted on every
dataset during the T6 audit, and that locus-disjoint generalization is not
established.

\item \textbf{Open access to data and code}

Question: Does the paper provide open access to the data and code, with
sufficient instructions to faithfully reproduce the main experimental results,
as described in supplemental material?

Answer: \answerNo{}

Justification: The evaluated datasets and model checkpoints are identified by
citations to their original public sources, and an anonymized code archive was
provided during review. However, the final public repository URL and the
availability of complete camera-ready T6, E3, E4, E8, and perturbation scripts,
configuration files, and reproducibility instructions have not yet been
verified. This answer can be updated to Yes only after the repository
and its contents are publicly accessible and checked.

\item \textbf{Experimental setting/details}

Question: Does the paper specify all the training and test details (e.g., data
splits, hyperparameters, how they were chosen, type of optimizer) necessary to
understand the results?

Answer: \answerYes{}

Justification: Section~3 describes the supplied held-out test partitions,
training-only tuning, model and feature choices, and the principal evaluation
protocol. The T6 Random Forest search uses 200 or 500 estimators, maximum
depth~20, square-root feature subsampling, minimum split size~2, and a
four-fold stratified training-set CV (seed~42), using AUROC for binary tasks and
accuracy for multiclass tasks; test MCC is reported after refitting. The
robustness experiments use their separately described probe/pooling and RF
configurations. The order-perturbation experiment uses a scaled logistic
regression with fixed $C=1$ and ridge $\alpha=1$, with frozen and retrained
probes. The precise protocols differ across the historical benchmark, E3, E4,
E8, and T6, as explicitly noted in the paper.

\item \textbf{Experiment statistical significance}

Question: Does the paper report error bars suitably and correctly defined or
other appropriate information about the statistical significance of the
experiments?

Answer: \answerYes{}

Justification: Main MCC outcomes use the provided held-out test partitions,
not five-fold test-set averages. Comparisons across models are paired at the
dataset level. The historical retrospective best-$k$ oracle analysis reports
two-sided Wilcoxon signed-rank tests with Benjamini--Hochberg correction and a
mixed-effects analysis (Appendix~C); these $p$-values are not attributed to the
fixed-$k=6$ baseline. The order-perturbation analysis averages three shuffle
seeds per dataset before computing dataset-paired contrasts and reports a
bootstrap confidence interval for the NTv3 versus DNABERT-2 comparison
(Section~4.4 and Appendix~A). Claims are appropriately limited where new
fixed-6-mer significance tests have not been reported.

\item \textbf{Experiments compute resources}

Question: For each experiment, does the paper provide sufficient information
on the computer resources (type of compute workers, memory, time of execution)
needed to reproduce the experiments?

Answer: \answerNo{}

Justification: Appendix~F reports descriptive GFLOP estimates per 500-bp
sequence for the subset of methods with available cost data. These are
externally supplied operation-count estimates, not measured wall-clock time,
peak memory, or hardware-normalized throughput. Verified hardware, CPU/GPU
allocation, peak memory, and elapsed runtime are not provided for every
experiment. Consequently, the previously stated single-A100 and approximately
20-GPU-hour total are not retained as verified resource measurements. This
answer should be revised only after a complete compute-resource audit.

\item \textbf{Code of ethics}

Question: Does the research conducted in the paper conform, in every respect,
with the NeurIPS Code of Ethics?

Answer: \answerYes{}

Justification: The experiments analyze previously released benchmark DNA
sequences and pretrained models, without recruiting participants, collecting
new personal genomic records, or generating subject-level predictions for
clinical use. Potential limitations and risks from overinterpreting frozen
representation performance are acknowledged, and the study does not release
new genomic sequences or pretrained biological models.

\item \textbf{Broader impacts}

Question: Does the paper discuss both potential positive societal impacts and
negative societal impacts of the work performed?

Answer: \answerYes{}

Justification: The Broader Impacts paragraph
in Section~5 discusses potential positive
impacts, including more reproducible genomic
model evaluation and reduced computational
expenditure through inexpensive compositional
baselines. It also addresses potential negative
impacts arising from overgeneralizing the
findings to fine-tuned models, clinical prediction,
or other genomic applications not evaluated
in this work.

\item \textbf{Safeguards}

Question: Does the paper describe safeguards that have been put in place for
responsible release of data or models that have a high risk for misuse?

Answer: \answerNA{}

Justification: The work does not train or release new sequence-generating
models, pretrained weights, or personal genomic datasets. It evaluates
previously available models on benchmark tasks and releases, or intends to
release, only experimental evaluation code and derived aggregate results.

\item \textbf{Licenses for existing assets}

Question: Are the creators or original owners of assets (e.g., code, data,
models), used in the paper, properly credited and are the license and terms of
use explicitly mentioned and properly respected?

Answer: \answerYes{}

Justification: The original benchmark publications and the five evaluated
model families are cited. The model repositories and their stated licenses are:
NTv3-650M (custom non-commercial license;
\url{https://huggingface.co/InstaDeepAI/NTv3_650M_pre});
HyenaDNA-160k (BSD-3-Clause;
\url{https://huggingface.co/LongSafari/hyenadna-medium-160k-seqlen-hf});
DNABERT-2 (Apache~2.0;
\url{https://huggingface.co/zhihan1996/DNABERT-2-117M});
Caduceus-Ph (Apache~2.0;
\url{https://huggingface.co/kuleshov-group/caduceus-ph_seqlen-131k_d_model-256_n_layer-16});
and Evo2-1B (Apache~2.0;
\url{https://huggingface.co/arcinstitute/evo2_1b_base}).
The work uses these assets for academic evaluation and does not redistribute
model weights. The authors should confirm the access terms for NTv3 and the
licenses and redistribution conditions of the underlying benchmark datasets
and any reused source code before submission.

\item \textbf{New assets}

Question: Are new assets introduced in the paper well documented and is the
documentation provided alongside the assets?

Answer: \answerNo{}

Justification: The paper does not introduce new training datasets, pretrained
foundation models, or model checkpoints. It does, however, develop evaluation
scripts and analyses that are relevant research artifacts. Their final public
packaging, documentation, and correspondence to the camera-ready results have
not yet been verified. Once the complete repository and reproducibility
instructions are released, this answer can be updated accordingly.

\item \textbf{Crowdsourcing and research with human subjects}

Question: For crowdsourcing experiments and research with human subjects, does
the paper include the full text of instructions given to participants and
screenshots, if applicable, as well as details about compensation (if any)?

Answer: \answerNA{}

Justification: No crowdsourcing studies or newly recruited human participants
were involved. The experiments use existing publicly released computational
benchmarks.

\item \textbf{Institutional review board (IRB) approvals or equivalent for
research with human subjects}

Question: Does the paper describe potential risks incurred by study
participants, whether such risks were disclosed to the subjects, and whether
Institutional Review Board (IRB) approvals (or an equivalent approval/review
based on the requirements of your country or institution) were obtained?

Answer: \answerNA{}

Justification: No prospective human-subject study or collection of identifiable
participant data was conducted as part of this work. The experiments use
existing benchmark resources, and an IRB review for new participant research is
not applicable to the activities reported here.

\item \textbf{Declaration of LLM usage}

Question: Does the paper describe the usage of LLMs if it is an important,
original, or non-standard component of the core methods in this research?

Answer: \answerYes{}

Justification: LLM tools were used to assist with language editing, manuscript
organization, and debugging selected implementation or plotting routines. They
were not used as biological prediction models, to generate benchmark labels, or
as part of the reported evaluation pipelines. The authors retain
responsibility for the experimental design, verification of analyses, and
scientific interpretation. No LLM constitutes a core methodological component
of the experiments.

\end{enumerate}